%% file: root.tex
\documentclass[letterpaper, 10 pt, conference]{ieeeconf}  

\IEEEoverridecommandlockouts                              

\usepackage{graphicx}       
\usepackage{booktabs}       
\usepackage{multirow}       
\usepackage[table]{xcolor}  
\usepackage{amsmath}        
\usepackage{amssymb}
\newif\ifnoavg
\noavgtrue          
\usepackage{subcaption}     

\usepackage{pifont}
\newcommand{\cmark}{\ding{51}}
\newcommand{\xmark}{\ding{55}}

\title{\LARGE \bf
FreeView: Robust 3D Reconstruction from Uncurated Views
}

\author{
Junho Kim$^{*}$, Hyunyoung Choi$^{*}$, and Seongwon Lee$^{\dagger}$%
\thanks{J.\,Kim, H.\,Choi, and S.\,Lee are with the School of Electrical Engineering, Kookmin University, Seoul 02707, South Korea, {\tt\small \{jhk00,kellychy,sungonce\}@kookmin.ac.kr}.}%
\thanks{$^{*}$ J.\,Kim and H.\,Choi contributed equally to this work.}%
\thanks{$^{\dagger}$ S.\,Lee is the corresponding author.}%
}

\makeatletter
\let\@oldmaketitle\@maketitle
\renewcommand{\@maketitle}{%
    \@oldmaketitle
    \centering
    \includegraphics[width=\textwidth]{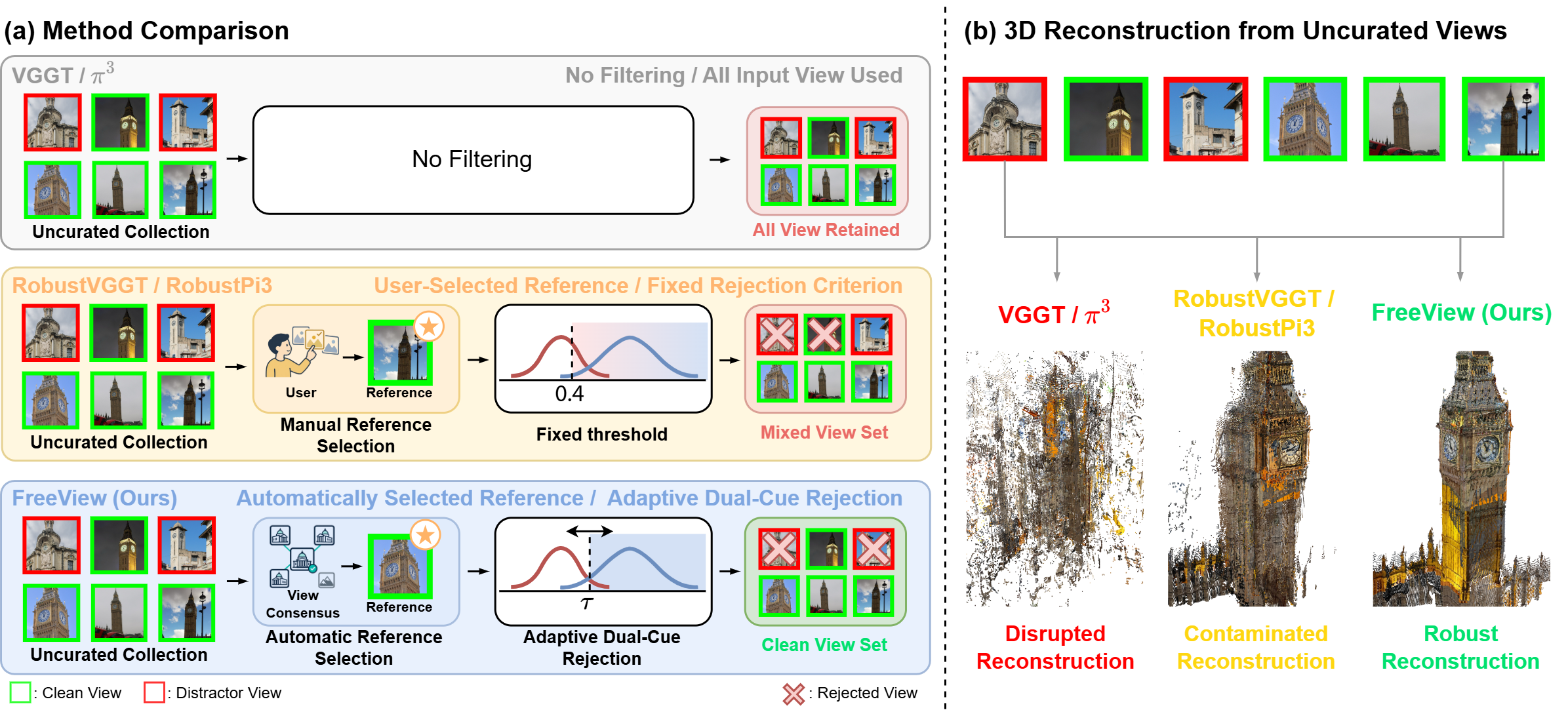}
    \captionof{figure}{\textbf{Robust 3D reconstruction from uncurated views.} VGGT~\cite{wang2025vggt} and $\pi^3$~\cite{wang2026pi3} process all input views without explicit distractor rejection. RobustVGGT and RobustPi3~\cite{han2026emergent} enable view filtering but rely on a clean reference and fixed rejection criteria. Our proposed FreeView automatically selects a representative reference and adapts rejection to each collection, preserving useful views while suppressing distractors without manual reference curation.}
    \label{fig:fig1}
    \setcounter{figure}{1}
}
\makeatother

\begin{document}

\maketitle
\thispagestyle{empty}
\pagestyle{empty}

\begin{abstract}

Reconstructing a coherent 3D scene from uncurated image collections remains challenging due to the presence of irrelevant or non-overlapping views. Recent work has shown that feed-forward reconstruction models contain internal view-consistency signals that support distractor rejection. However, reference-conditioned filtering relies on a curated clean reference, which is not guaranteed in practical image collections. To address this limitation, we introduce FreeView, a training-free framework that jointly selects a filtering reference and rejects distractor views from an uncurated collection. FreeView leverages collection-level consensus to identify a representative view that serves as a reliable reference for filtering. It then adapts view rejection to each input collection by combining complementary visual and geometric evidence. This approach preserves useful views under substantial appearance and viewpoint variations while rejecting distractors. FreeView integrates with both VGGT and $\pi^3$, removing the need for a manually curated reference or manual calibration of view rejection. Extensive experiments across multiple datasets demonstrate high clean-view retention and distractor rejection rates, leading to reliable 3D reconstruction from uncurated views.

\end{abstract}

\section{INTRODUCTION}
Recovering coherent 3D geometry from visual observations is a fundamental problem in robotic perception and visual mapping. Multi-view reconstruction~\cite{Schonberger_2016_CVPR, pan2024glomap, he2024detectorfree, lee2025densesfm, SSS:2006} provides a representation of the surrounding environment by jointly recovering scene structure and camera poses. In practical settings, however, images collected across different capture sessions or retrieved from large databases may contain observations of unrelated scenes. Such collections require a reconstruction system to determine which views support a common scene, while excluding distractors that can compromise the recovered geometry.

Recent advances in feed-forward reconstruction~\cite{yang2025fast3r,cabon2025must3r,wang2025cut3r, tang2024mv, wang2026amb3r} have enabled direct prediction of camera poses and scene geometry from multiple images. The Visual Geometry Grounded Transformer (VGGT)~\cite{wang2025vggt} jointly predicts camera parameters and dense scene geometry from multiple input views, while $\pi^3$~\cite{wang2026pi3} introduces permutation-equivariant geometric prediction without a fixed reference view. These models provide efficient foundations for multi-view scene understanding and motivate the use of feed-forward reconstruction beyond carefully prepared image sets. Their application to uncurated collections, however, also requires mechanisms for identifying images that should not participate in the reconstruction.

RobustVGGT~\cite{han2026emergent} addresses this challenge by exploiting the view-consistency signals encoded in VGGT's internal representations, with its RobustPi3 variants extending the approach to $\pi^3$. These signals provide evidence of a view's relevance to a given reference, enabling distractor rejection without additional training. This finding demonstrates that pretrained geometric models contain useful information for deciding which observations belong together. However, applying reference-conditioned filtering to uncurated collections requires establishing a trustworthy reference without prior knowledge of which input views belong to the target scene.

The filtering reference plays a central role because it determines the scene against which the remaining views are assessed. When a distractor is used as this reference, clean views may appear inconsistent with it, while images of an unrelated scene may receive stronger support. Consequently, an effective relevance signal can still produce an unsuitable selection if it is conditioned on the wrong reference. This dependence arises in view filtering, even when the reconstruction model does not require a fixed reference frame. Thus, $\pi^3$’s permutation-equivariant geometry prediction does not eliminate the need for a reliable reference when filtering views. In an uncurated collection, this reference must itself be inferred from the relationships among the inputs.

A second challenge is balancing distractor rejection with the preservation of useful scene observations. Clean views can differ substantially in appearance and viewpoint, and their relevance may not be equally evident from every visual or geometric cue. An overly restrictive decision can discard valid observations and reduce scene coverage, whereas a permissive decision can retain distractors that degrade reconstruction. Moreover, the degree of agreement among views varies across collections, and different reconstruction backbones can yield different relevance-score distributions. These considerations motivate a rejection strategy that accounts for the evidence available in each input set while preserving views that remain useful for reconstruction.

In this paper, we introduce FreeView, a training-free framework for robust 3D reconstruction without a pre-verified reference view. FreeView addresses reference selection and view rejection within a unified inference pipeline that can be paired with either VGGT or $\pi^3$. We focus on collections in which the target scene forms a coherent group with stronger local visual support than competing distractor groups. Our approach uses consensus within the collection to establish a representative reference for filtering and adapts the subsequent rejection process to the available visual and geometric evidence.

First, we introduce Consensus Reference Selection, which identifies a representative view through the strength of its agreement with other images. The underlying intuition is that clean views tend to reinforce one another through shared scene content, whereas unrelated distractors receive weaker support. By assessing this support before reference-conditioned filtering, FreeView establishes a basis for interpreting the remaining views without relying on a designated input position. This makes reference selection an explicit part of processing uncurated inputs and removes the need to supply a clean image in advance.

Second, we introduce Adaptive Dual-Cue Rejection, which combines visual evidence with geometry-aware evidence from the selected reconstruction backbone to determine which views should be retained. The rejection process adapts to each collection, accommodating differences in appearance, viewpoint, and cross-view agreement.Views supported by either cue are retained, while those unsupported by both are rejected. This allows useful observations to survive filtering even when their relevance is not equally apparent from both cues. The selected views are then passed to the same backbone for final reconstruction. Together, the two components support both VGGT and $\pi^3$ without additional model training or manual calibration of the rejection process.

Experiments across multiple datasets demonstrate high clean-view retention and distractor rejection rates without manual reference selection. These results show that addressing reference selection together with adaptive rejection enables reliable 3D reconstruction from uncurated image collections.

Our main contributions can be summarized as follows:
\begin{itemize}
    \item We present FreeView, a training-free view-curation framework for robust 3D reconstruction that integrates with both VGGT and $\pi^3$. Its Consensus Reference Selection establishes a representative filtering reference directly from the input collection, removing the need for a pre-verified clean image.
    \item We introduce Adaptive Dual-Cue Rejection, which adapts view selection to each collection using complementary visual and backbone-derived geometric evidence, addressing the balance between clean-view preservation and distractor rejection without manual calibration.
    \item We instantiate FreeView with two reconstruction backbones and evaluate view selection and reconstruction across multiple datasets. Our experiments demonstrate high clean-view retention and distractor rejection rates, enabling reliable reconstruction across varying distractor levels and reference conditions.
\end{itemize}

\section{Related Work}

\subsection{Feed-Forward 3D Reconstruction}

Recent learning-based approaches~\cite{yang2025fast3r,cabon2025must3r,wang2025cut3r} have substantially simplified multi-view 3D reconstruction through learned geometric prediction.
DUSt3R~\cite{wang2024dust3r} introduced pointmap regression from image pairs, while MASt3R~\cite{leroy2024mast3r} strengthened learned correspondences and MASt3R-SfM~\cite{duisterhof2025mast3rsfm} incorporated these representations into a scalable reconstruction pipeline.
VGGSfM~\cite{wang2024vggsfm} combined learned visual geometry with classical structure-from-motion components, while VGGT~\cite{wang2025vggt} unified camera, depth, and pointmap prediction within a transformer-based feed-forward architecture.
$\pi^3$~\cite{wang2026pi3} introduced permutation-equivariant geometric reasoning, removing dependence on a fixed reference view in geometry prediction.

These methods primarily focus on improving geometric prediction and reconstruction architectures.
Our work addresses the complementary problem of input-set robustness: given an uncurated collection containing unrelated views, we aim to identify a coherent subset before final reconstruction.
In particular, permutation equivariance addresses changes in input ordering, while view curation determines which observations should participate in reconstruction.
FreeView provides this curation through a common framework that integrates with both VGGT and $\pi^3$.

\begin{figure*}
  \centering
  \includegraphics[width=1.0\textwidth]{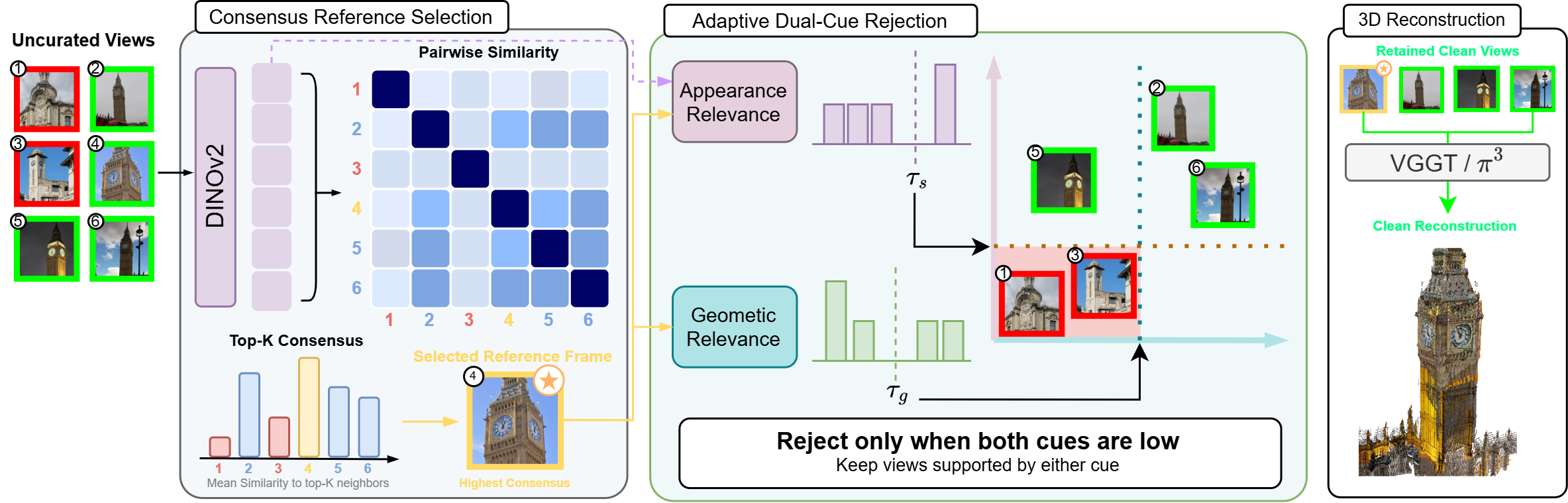}
  \caption{\textbf{Overview of FreeView. }Given an uncurated image collection, Consensus Reference Selection identifies a representative view through agreement among related images, establishing a reference for filtering without prior knowledge of which views are clean. Adaptive Dual-Cue Rejection combines complementary visual and geometric evidence to adapt view selection to each collection. Retaining views supported by either cue preserves useful observations under appearance and viewpoint variations, while rejecting views unsupported by both suppresses distractors. The retained views are then processed by VGGT or $\pi^3$ for final reconstruction.}
  \label{fig:fig2}
\end{figure*}

\subsection{Robust View Filtering for 3D Reconstruction}

Robustness to erroneous observations has long been an integral part of classical structure-from-motion pipelines~\cite{Lindenberger_2021_ICCV}.
Conventional systems use image retrieval, local feature matching, geometric verification, robust estimation~\cite{Barath_2020_CVPR}, and view-graph optimization to prevent inconsistent observations from corrupting reconstruction.
Although effective, these procedures rely on multiple stages, many of which recent feed-forward reconstruction models bypass.

RobustVGGT~\cite{han2026emergent} demonstrated that VGGT implicitly encodes view-relevance information in its internal attention and feature representations.
Its RobustPi3 variants further explored the use of analogous signals in $\pi^3$.
By measuring a view's relevance to a reference, distractors can be rejected without additional training.
However, this filtering process is conditioned on a predefined reference view.
When the reference itself is unrelated to the target scene, the resulting relevance scores can favor an unsuitable subset.
In addition, a fixed rejection criterion may not adapt well to collection-specific variations in appearance, viewpoint, and view consistency.

FreeView addresses these limitations by treating reference selection as part of the robust reconstruction problem itself.
We first infer a representative anchor from collection-level consensus and then perform adaptive view rejection using complementary visual and geometry-aware cues.
The geometric evidence is obtained from the selected reconstruction backbone, allowing the same reference-selection and rejection framework to be used with VGGT~\cite{wang2025vggt} and $\pi^3$~\cite{wang2026pi3}.
This removes the need for a pre-verified clean reference and manual calibration of the rejection process.

\subsection{Reference Selection and Visual Retrieval}

Selecting representative or relevant images is closely related to visual place recognition and image retrieval~\cite{Arandjelovic_2016_CVPR}.
Classical retrieval methods and recent learned systems such as MegaLoc~\cite{Berton_2025_MegaLoc} provide global image representations for identifying visually related views~\cite{Arandjelovic_2016_CVPR, 9879390, 10204303, Izquierdo_2024_CVPR}.
Self-supervised visual representations such as DINOv2~\cite{oquab2024dinov2} and DINOv3~\cite{simeoni2025dinov3} likewise provide descriptors that capture visual and semantic similarity.

While visual retrieval primarily aims to identify images related to a query, our goal is to establish a reliable filtering reference from a collection containing both relevant and unrelated views.
FreeView uses pairwise DINOv2 similarities to estimate collection-level consensus and selects an anchor supported by a coherent group of views.
FreeView thus uses visual similarity both to establish the filtering reference and to support view retention alongside geometry-aware evidence.

\section{Methodology}

\subsection{Preliminaries}

\textbf{Feed-forward 3D reconstruction.}
Given a set of $N$ input images
$\mathcal{I}=\{I_i\}_{i=1}^{N}$,
we consider two representative feed-forward reconstruction models,
VGGT~\cite{wang2025vggt} and $\pi^3$~\cite{wang2026pi3}.
They jointly predict
\begin{equation}
    \left\{
    (\mathcal{P}_i,\mathrm{D}_i,\mathrm{X}_i,\mathrm{C}_i)
    \right\}_{i=1}^{N},
\end{equation}
where $\mathcal{P}_i$ denotes the camera parameters,
$\mathrm{D}_i$ the per-pixel depth map,
$\mathrm{X}_i$ the 3D pointmap, and
$\mathrm{C}_i$ the corresponding confidence map.
VGGT anchors its reconstruction to the coordinate system of the first
input camera, making the first view a reconstruction reference.
In contrast, $\pi^3$ adopts permutation-equivariant multi-view reasoning
and represents geometry in relative coordinate systems without relying
on the first input view as the reconstruction reference. Both models
encode cross-view geometric information in their internal representations
that can be exploited for view relevance estimation.

\textbf{Reference-conditioned filtering.}
RobustVGGT~\cite{han2026emergent} and its $\pi^3$ variants designate one input
view as a filtering reference.
Given a reference image $I_r$, they measure the relevance of each
remaining view $I_j$ to $I_r$ using internal representations of the
corresponding reconstruction backbone and reject views whose relevance
falls below a fixed threshold.
Accordingly, the retained view set is
\begin{equation}
    \mathcal{V}_r
    =
    \left\{
        I_j \in \mathcal{I}
        \mid
        s(I_r,I_j) \geq \tau
    \right\},
\end{equation}
where $s$ denotes a reference-conditioned relevance score,
$\tau$ is a fixed rejection threshold, and
$\mathcal{V}_r$ denotes the subset retained with respect to $I_r$.
Thus, regardless of the reconstruction backbone, the retained set
$\mathcal{V}_r$ depends on the choice of filtering reference $I_r$.

\subsection{Problem Formulation}

We consider an uncurated collection of $N$ images,
\begin{equation}
    \mathcal{I}=\mathcal{C}\cup\mathcal{D},
    \qquad
    \mathcal{C}\cap\mathcal{D}=\emptyset,
\end{equation}
where $\mathcal{C}$ contains views of the target scene and $\mathcal{D}$ contains distractor views. The membership of each image is unknown at inference time, and no clean reference is supplied.

We assume that the target scene forms a coherent group with stronger local visual support than competing distractor groups. This assumption does not require target views to constitute a majority of the collection.

Our goal is to select a filtering reference $I_r\in\mathcal{C}$ and aims to recover a subset ($\hat{\mathcal{C}}\subseteq\mathcal{C}$ that retains target-scene views while excluding distractors. The retained subset is then passed to a feed-forward reconstruction backbone to estimate scene geometry and camera poses.

\subsection{Consensus Reference Selection}

Reference-conditioned filtering requires a reference that reliably
represents the target scene. Since no input view is known a priori to provide such a reference, we infer one from the mutual agreement among the views.

For each image $I_i$, we extract its DINOv2~\cite{oquab2024dinov2} CLS token
$\mathbf{z}_i^{\mathrm{cls}}$ and $\ell_2$-normalize it as
$\mathbf{e}_i=\mathbf{z}_i^{\mathrm{cls}}/
\|\mathbf{z}_i^{\mathrm{cls}}\|_2$.
We then compute the pairwise appearance similarity
\begin{equation}
    S_{ij}^{a}
    =
    \mathbf{e}_i^\top \mathbf{e}_j .
\end{equation}

Rather than measuring agreement with the entire collection, we evaluate
the local support of each view using its $K$ most similar neighbors.
Let $\mathcal{N}_K(i)$ denote the indices of the top-$K$ similarities
to $I_i$, excluding itself. We define the consensus score as
\begin{equation}
    c_i
    =
    \frac{1}{K}
    \sum_{j\in\mathcal{N}_K(i)}
    S_{ij}^{a},
\end{equation}
and select the filtering reference by
\begin{equation}
    i_r
    =
    \arg\max_i c_i,
    \qquad
    I_r=I_{i_r}.
\end{equation}

\begin{figure}
  \centering
  \includegraphics[width=0.48\textwidth]{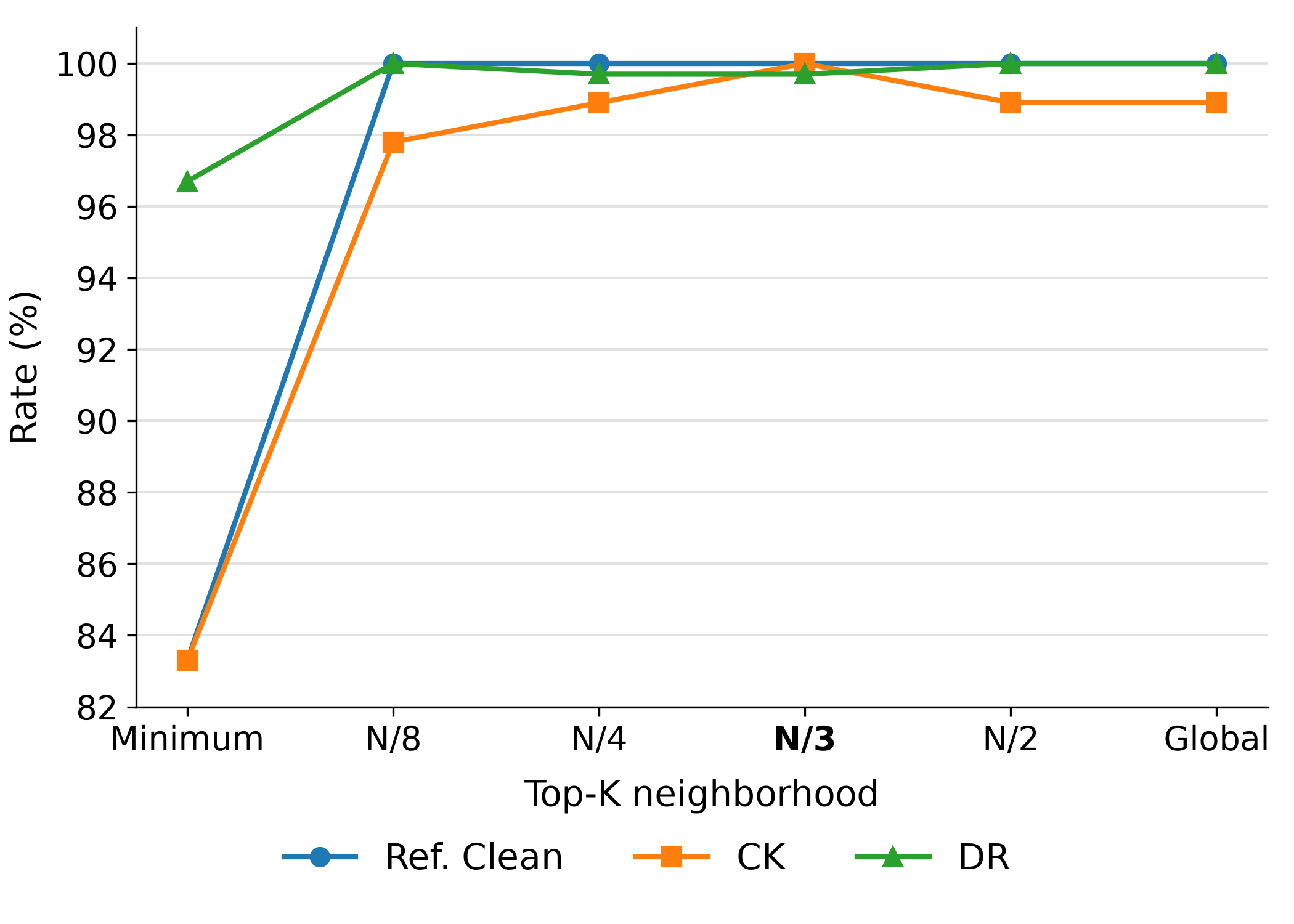}
  \caption{{Effect of neighborhood size on reference selection and view filtering on Phototourism under the Large setting, with 30 clean views and 50 distractors.}}
  \vspace{-4mm}
  \label{fig:topk}
\end{figure}

This local consensus favors views supported by a coherent subset without
requiring the target views to dominate the entire collection.
We use $K=\max(3,\lfloor N/3\rfloor)$ for all experiments and analyze
its sensitivity in Fig.~\ref{fig:topk}.

After selecting $I_r$, we obtain the backbone representations used for
geometry-aware filtering.
For VGGT~\cite{wang2025vggt}, whose aggregated representations depend on the designated
first view, we move $I_r$ to the first input position and recompute the
multi-view features.
For $\pi^3$~\cite{wang2026pi3}, permutation equivariance makes such input reordering
unnecessary.
Because the appearance descriptors are extracted independently for each image, their pairwise similarities $S_{ij}^{a}$ remain unchanged by input reordering.

\subsection{Adaptive Dual-Cue Rejection}

Given the selected filtering reference $I_r$, we estimate the relevance
of each remaining view using complementary appearance and geometry-aware
signals. Rather than applying a fixed rejection threshold, we adapt the
decision boundary to the score distribution of each input collection.

\textbf{Appearance relevance.}
We directly reuse the DINOv2 similarity from the reference-selection
stage. The appearance relevance of $I_j$ is
\begin{equation}
    r_j^a
    =
    S_{i_r,j}^{a}.
\end{equation}

\textbf{Geometry-aware relevance.}
Let $\mathbf{f}_{i,t}\in\mathbb{R}^{d}$ denote the $t$-th spatial
feature of image $I_i$ extracted after multi-view aggregation in the
selected reconstruction backbone.
We L2-normalize each spatial feature, average the normalized features over the $T$ spatial tokens, and L2-normalize the resulting image-level descriptor:
\begin{equation}
    \mathbf{g}_i
    =
    \frac{
        \frac{1}{T}\sum_{t=1}^{T}
        \frac{\mathbf{f}_{i,t}}{\|\mathbf{f}_{i,t}\|_2}
    }{
        \left\|
        \frac{1}{T}\sum_{t=1}^{T}
        \frac{\mathbf{f}_{i,t}}{\|\mathbf{f}_{i,t}\|_2}
        \right\|_2
    }.
\end{equation}
The corresponding geometry-aware relevance is
\begin{equation}
    r_j^g
    =
    \mathbf{g}_{i_r}^{\top}\mathbf{g}_j .
\end{equation}

Following the observation of RobustVGGT~\cite{han2026emergent} that
late-layer representations implicitly encode cross-view geometric
consistency, we use these backbone-derived features as a complementary
signal to the image-wise DINOv2 appearance descriptor.

\input{table/pose_eval_all}

\textbf{Adaptive view rejection.}
The distributions of appearance and geometry relevance vary across
image collections, making a fixed rejection threshold difficult to apply consistently across collections.
We therefore determine $\tau_a$ and $\tau_g$ independently for each
collection using Otsu's criterion~\cite{otsu1979threshold}, which partitions the relevance scores
by maximizing the separation between low- and high-relevance groups:
\begin{equation}
    \tau_a = \operatorname{Otsu}\left(\{r_j^a\}\right),
    \qquad
    \tau_g = \operatorname{Otsu}\left(\{r_j^g\}\right).
\end{equation}

Because Otsu's criterion always induces a partition even when the score
distribution is weakly separated, we suppress rejection from a cue when
the mean separation between the resulting groups is below
$\gamma=0.04$, which is fixed across all experiments.

To preserve valid views when the two cues disagree, we adopt a
conservative rule and reject a view only when both relevance scores fall
below their respective thresholds:
\begin{equation}
    \operatorname{Reject}(I_j)
    =
    [r_j^a < \tau_a]
    \land
    [r_j^g < \tau_g].
    \label{eq:dual_cue_rejection}
\end{equation}
The retained set is therefore
\begin{equation}
    \hat{\mathcal{C}}
    =
    \left\{
        I_j \in \mathcal{I}
        \mid
        r_j^a \geq \tau_a
        \ \lor\
        r_j^g \geq \tau_g
    \right\}.
\end{equation}

We apply this rejection rule for two rounds.
The selected reference $I_r$ remains fixed, while $\tau_a$ and $\tau_g$
are recomputed over the currently retained views at each round.
The final retained set is then passed to the selected reconstruction
backbone for reconstruction.

\input{ablation/module}

\input{ablation/filtering}

\section{Experiments}

\subsection{Experimental Setup}

\textbf{Datasets and uncurated benchmarks.}
We evaluate FreeView on three multi-view reconstruction datasets:
Phototourism~\cite{SSS:2006}, On-the-Go~\cite{Ren2024NeRF}, and ETH3D~\cite{schops2017eth3d}. For each dataset, we construct controlled uncurated collections by
combining views of a target scene with unrelated views sampled from
other scenes.
For Phototourism and On-the-Go, each collection contains
$N_c=30$ clean views and
$N_d\in\{10,30,50\}$ distractors for the
\emph{Small}, \emph{Medium}, and \emph{Large} settings, respectively,
resulting in $40$, $60$, and $80$ input views.
For ETH3D, where fewer views are available per scene, we use
$N_c=14$ clean views and
$N_d\in\{5,14,30\}$ distractors, resulting in
$19$, $28$, and $44$ input views.
The clean subset is fixed across noise levels, while the distractors
are progressively added such that
$\mathcal{D}_{\mathrm{Small}}
\subset
\mathcal{D}_{\mathrm{Medium}}
\subset
\mathcal{D}_{\mathrm{Large}}$.

\textbf{Evaluation protocol.}
We evaluate four settings: \emph{Clean-only}, \emph{Clean Reference},
\emph{Distractor Reference}, and automatic reference selection with
FreeView.
Clean-only uses only target-scene views and serves as an oracle setting.
For reference-conditioned methods, Clean Reference and Distractor
Reference use the same mixed collection and differ only in whether the
filtering reference is a target-scene view or a distractor.
We evaluate five paired reference choices per target scene.
FreeView receives the same mixed collection without a designated
reference and selects one automatically.
For fixed-threshold retrieval baselines (MegaLoc~\cite{Berton_2025_MegaLoc} and DINO~\cite{oquab2024dinov2, simeoni2025dinov3}), we tune the
threshold per dataset on the Clean Reference setting by maximizing the
mean of CK and DR, and keep it fixed for Distractor Reference evaluation.

\textbf{Evaluation metrics.}
We evaluate filtering using the clean-view keep rate (CK) and distractor
rejection rate (DR), defined as
$\mathrm{CK}=|\hat{\mathcal{C}}\cap\mathcal{C}|/|\mathcal{C}|$
and
$\mathrm{DR}=|\mathcal{D}\setminus\hat{\mathcal{C}}|/|\mathcal{D}|$,
respectively. Severe rejection of clean views can leave too few camera poses for valid alignment, preventing meaningful ATE evaluation. Even when alignment is possible, ATE and RPE computed only on the retained views can favor methods that discard difficult but valid observations. We therefore report coverage-aware pose recall over all clean ground-truth views, counting rejected views as failures.
The retained clean camera poses are aligned to the ground truth using a Umeyama alignment~\cite{umeyama1991least}.
Translation error is normalized by the scene radius
$\rho=\sqrt{|\mathcal{C}|^{-1}\sum_{i\in\mathcal{C}}
\|\mathbf{c}_i-\bar{\mathbf{c}}\|_2^2}$,
and $R_t$ is reported at $2\%$, $5\%$, and $10\%$ of $\rho$.
Rotation recall $R_r$ is computed from the angular error of the aligned
camera orientations at thresholds of $2^\circ$, $5^\circ$, and
$10^\circ$.
Rejected clean views are counted as failures for both metrics.

\begin{figure*}
  \centering
  \includegraphics[width=1.0\textwidth]{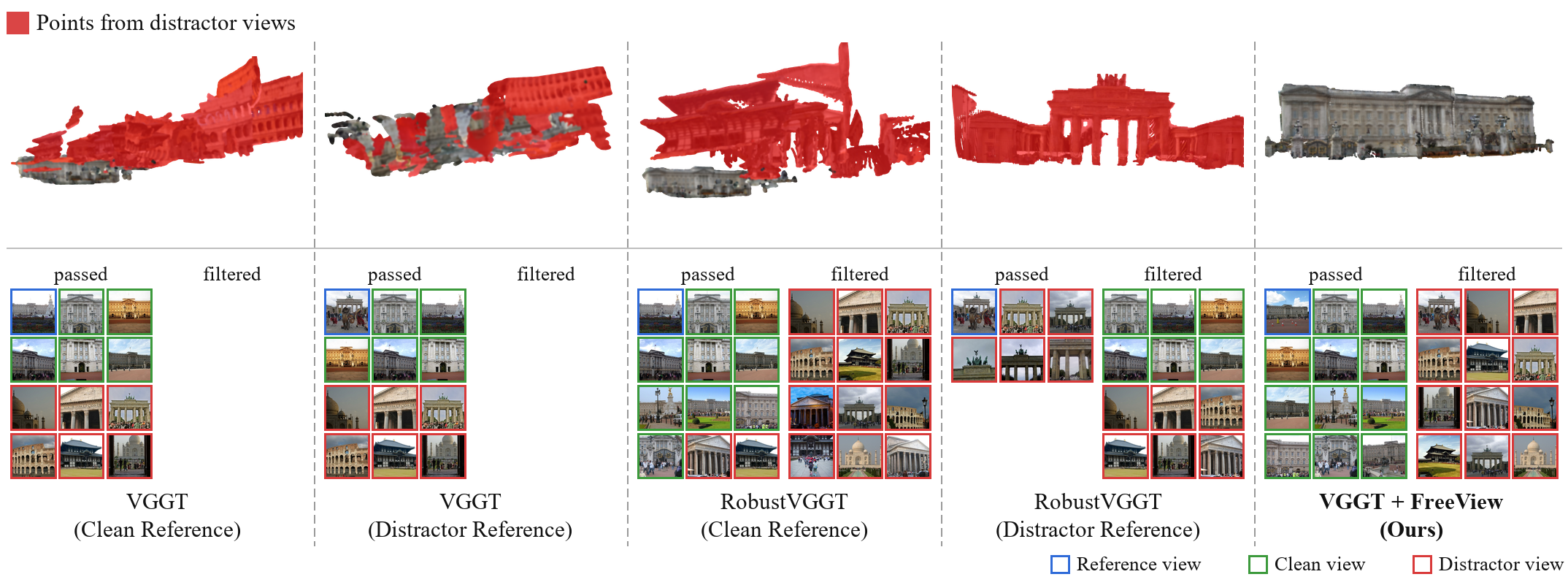}
  \caption{Qualitative comparison of view filtering and 3D reconstruction.}
  \label{fig:qualitative}
  \vspace{-5mm}
\end{figure*}

\subsection{Experimental Results}

\textbf{Camera pose estimation.}
Quantitative results are summarized in Table~\ref{tab:pose_eval_all_avg}.

(i) \emph{VGGT-based methods.}
With a clean filtering reference, retrieval-based methods and RobustVGGT
effectively remove distractors and maintain strong pose recall, showing
that reference-conditioned relevance is reliable when the reference
belongs to the target scene.
However, when the same methods are given a distractor reference,
clean-view retention and pose recall collapse across all datasets,
despite using the same underlying image collections.
This demonstrates that strong visual or geometry-aware relevance alone
is insufficient when conditioned on an unreliable reference.
FreeView avoids this failure by automatically selecting the
filtering reference, achieving consistently strong filtering and camera
reconstruction across all noise levels.

(ii) \emph{$\pi^3$-based methods.}
Although $\pi^3$ itself is permutation-equivariant, RobustPi3 still
depends on a designated filtering reference and degrades substantially
when a distractor is used as the reference.
The consistent improvement on $\pi^3$ indicates that the benefits of
FreeView are not specific to VGGT's first-view-anchored reconstruction.

(iii) \emph{Robustness to distractors.}
Across Phototourism, ETH3D, and On-the-Go, both FreeView variants remain
robust as the distractor level increases from Small to Large, while
avoiding the catastrophic failures observed with incorrect references.
Overall, FreeView provides consistent reference selection and view
filtering across both VGGT and $\pi^3$.

\textbf{Qualitative results.}
Figure~\ref{fig:qualitative} illustrates the effect of reference selection
on view filtering and reconstruction.
VGGT retains distractor observations, resulting in spurious geometry,
while RobustVGGT substantially improves filtering with a clean reference,
although residual views from an unrelated landmark still contaminate the
reconstruction.
With a distractor reference, RobustVGGT instead rejects valid target views
and produces a degraded reconstruction.
FreeView automatically selects a target-consistent reference, preserving
clean views while suppressing distractors and recovering a coherent
reconstruction without manual reference selection.

\subsection{Ablation and Analysis}

We conduct all ablations under the Large setting to examine the design choices at the highest distractor level in our benchmark.

\textbf{Model component design.}
Table~\ref{tab:component_ablation} examines the contribution of Consensus Reference Selection and Adaptive Dual-Cue Rejection under the Large-noise setting with a distractor reference.
Adding reference selection alone yields a substantial performance recovery across all three datasets, highlighting the effectiveness of collection-level consensus for establishing a reliable filtering anchor. Adaptive rejection then adjusts the rejection criterion to the score
distribution of each collection, yielding dataset-dependent improvements
in filtering and pose reconstruction.
Together, the two components provide robust view selection across
collections with different visual and geometric characteristics.

\textbf{Top-$K$ sensitivity.}
Figure~\ref{fig:topk} studies the effect of the neighborhood size used
for consensus reference selection on Phototourism under the Large-noise
setting.
Using only the minimum number of neighbors leads to noticeably lower
reference-selection accuracy and clean-view retention, indicating that
overly local consensus is sensitive to small accidental clusters.
Performance improves substantially as the neighborhood expands and
remains stable over a broad intermediate range.
Among these settings, $K=\max(3,\lfloor N/3\rfloor)$ provides the most
consistent overall performance across reference selection and view filtering.
We therefore use $N/3$ as the default setting in all experiments.

\textbf{Dual-cue fusion filtering.}
Table~\ref{tab:cue_ablation} compares the individual relevance
cues and alternative fusion strategies under the Large-noise setting.
The relative effectiveness of the two cues varies across datasets:
DINOv2 appearance relevance performs particularly well on Phototourism
and On-the-Go, while the geometry-aware cue remains competitive on ETH3D metrics.
These results indicate that the two cues capture different aspects of
view relevance.
We therefore use a conservative AND rejection rule, removing a view
only when both cues indicate low relevance.
This design achieves consistently high clean-view retention while
maintaining strong distractor rejection across all datasets.
In contrast, OR rejection and direct score averaging are more aggressive
and frequently discard valid target-scene views, reducing scene
coverage.

\section{CONCLUSIONS}

In this paper, we presented FreeView, a training-free framework for robust 3D reconstruction from uncurated image collections. FreeView combines Consensus Reference Selection with Adaptive Dual-Cue Rejection to establish a representative filtering reference and preserve useful views through complementary visual and geometric evidence. This design removes the need for a pre-verified clean reference and manual calibration of view rejection. By integrating with both VGGT and $\pi^3$, FreeView provides a common view-curation framework for different reconstruction backbones. Experiments across multiple datasets demonstrate high clean-view retention and distractor rejection rates, enabling reliable reconstruction across varying distractor levels and
reference conditions. These results highlight the importance of jointly addressing reference selection and adaptive rejection when extending feed-forward reconstruction to uncurated inputs.

\bibliographystyle{IEEEtran}
\bibliography{ref}
\end{document}

%% file: table/pose_eval_all.tex

\begin{table*}[p]
    \centering
    \captionsetup{font=small,skip=3pt}
    \captionsetup[subtable]{font=small,skip=2pt,position=top}
    \caption{\textbf{View filtering and pose estimation across distractor levels.} For the same mixed image collections, reference-conditioned methods are evaluated with either a clean or a distractor reference to assess sensitivity to reference choice, while FreeView selects its reference automatically. CK/DR denote clean-view retention/distractor rejection; T/R denote translation/rotation recall over all clean views, counting rejected views as failures. For $\pi^3$, the reference is used only for filtering. \textbf{Bold}/\underline{underline} indicate the best/second-best non-oracle results per backbone. $^\dagger$ denotes DINOv2 features from the VGGT encoder.}
    \label{tab:pose_eval_all_avg}
    \setlength{\tabcolsep}{3pt}
    \renewcommand{\arraystretch}{1.0}

    \begin{subtable}{\textwidth}
        \centering
        \caption{Phototourism}
        \label{tab:pose_eval_pt}
        \resizebox{\linewidth}{!}{%
\begin{tabular}{l|cccc|cccc|cccc|cccc}
        \toprule
        \multirow{2}{*}{Methods}
        & \multicolumn{4}{c|}{Small} 
        & \multicolumn{4}{c|}{Medium} 
        & \multicolumn{4}{c|}{Large} 
        & \multicolumn{4}{c}{Avg} \\

        & CK$\uparrow$
        & DR$\uparrow$
        & T@2\,/\,5\,/\,10\%$\uparrow$
        & R@2\,/\,5\,/\,10$^\circ$$\uparrow$
        & CK$\uparrow$
        & DR$\uparrow$
        & T@2\,/\,5\,/\,10\%$\uparrow$
        & R@2\,/\,5\,/\,10$^\circ$$\uparrow$
        & CK$\uparrow$
        & DR$\uparrow$
        & T@2\,/\,5\,/\,10\%$\uparrow$
        & R@2\,/\,5\,/\,10$^\circ$$\uparrow$
        & CK$\uparrow$
        & DR$\uparrow$
        & T@2\,/\,5\,/\,10\%$\uparrow$
        & R@2\,/\,5\,/\,10$^\circ$$\uparrow$ \\
        \midrule
\rowcolor{gray!20}
\multicolumn{17}{c}{\textbf{Oracle (Using Clean Images Only; No Distractors)}} \\

        \rowcolor{gray!10}
        VGGT (Clean-only)~\cite{wang2025vggt}
        & 100.0 & -- & 49.2\,/\,79.2\,/\,93.8 & 96.7\,/\,99.9\,/\,100.0
        & 100.0 & -- & 49.2\,/\,79.2\,/\,93.8 & 96.7\,/\,99.9\,/\,100.0
        & 100.0 & -- & 49.2\,/\,79.2\,/\,93.8 & 96.7\,/\,99.9\,/\,100.0
        & 100.0 & -- & 49.2\,/\,79.2\,/\,93.8 & 96.7\,/\,99.9\,/\,100.0 \\

        \rowcolor{gray!10}
        $\pi^3$ (Clean-only)~\cite{wang2026pi3}
        & 100.0 & -- & 10.2\,/\,52.0\,/\,79.7 & 61.2\,/\,90.1\,/\,98.7
        & 100.0 & -- & 10.2\,/\,52.0\,/\,79.7 & 61.2\,/\,90.1\,/\,98.7
        & 100.0 & -- & 10.2\,/\,52.0\,/\,79.7 & 61.2\,/\,90.1\,/\,98.7
        & 100.0 & -- & 10.2\,/\,52.0\,/\,79.7 & 61.2\,/\,90.1\,/\,98.7 \\

        \rowcolor{gray!20}
        \multicolumn{17}{c}{\textbf{Using Clean Image as Reference}} \\

        VGGT~\cite{wang2025vggt}
        & \textbf{100.0} & 0.0
        & 42.0\,/\,73.1\,/\,88.6
        & 88.4\,/\,97.7\,/\,\underline{99.9}
        & \textbf{100.0} & 0.0
        & 36.4\,/\,72.1\,/\,86.3
        & 85.4\,/\,95.8\,/\,\underline{99.7}
        & \textbf{100.0} & 0.0
        & 32.2\,/\,70.2\,/\,85.1
        & 85.7\,/\,97.4\,/\,\underline{99.9}
        & \textbf{100.0} & 0.0
        & 36.9\,/\,71.8\,/\,86.7
        & 86.5\,/\,97.0\,/\,\underline{99.8} \\

        $\pi^3$~\cite{wang2026pi3}
        & \textbf{100.0} & 0.0
        & 10.9\,/\,43.2\,/\,75.4
        & 53.6\,/\,\underline{92.6}\,/\,\textbf{98.3}
        & \textbf{100.0} & 0.0
        & \underline{9.7}\,/\,35.3\,/\,60.0
        & 44.4\,/\,\underline{86.0}\,/\,96.0
        & \textbf{100.0} & 0.0
        & 5.8\,/\,29.2\,/\,54.7
        & 35.7\,/\,73.8\,/\,85.0
        & \textbf{100.0} & 0.0
        & 8.8\,/\,35.9\,/\,63.4
        & 44.6\,/\,\underline{84.1}\,/\,93.1 \\

        MegaLoc~\cite{Berton_2025_MegaLoc}+VGGT~\cite{wang2025vggt}
        & \textbf{100.0} & \textbf{100.0}
        & \underline{49.2}\,/\,\textbf{79.2}\,/\,\underline{93.8}
        & \underline{96.7}\,/\,\underline{99.9}\,/\,\textbf{100.0}
        & \textbf{100.0} & \underline{99.9}
        & 49.4\,/\,\underline{79.2}\,/\,\underline{93.8}
        & \underline{96.7}\,/\,\underline{99.9}\,/\,\textbf{100.0}
        & \textbf{100.0} & \underline{99.9}
        & \underline{49.6}\,/\,\underline{79.2}\,/\,\textbf{93.8}
        & \underline{96.7}\,/\,\underline{99.9}\,/\,\textbf{100.0}
        & \textbf{100.0} & \underline{99.9}
        & \underline{49.4}\,/\,\underline{79.2}\,/\,\underline{93.8}
        & \underline{96.7}\,/\,\underline{99.9}\,/\,\textbf{100.0} \\

        DINOv2$^\dagger$\cite{oquab2024dinov2}+VGGT~\cite{wang2025vggt}
        & \textbf{100.0} & \textbf{100.0}
        & \underline{49.2}\,/\,\textbf{79.2}\,/\,\underline{93.8}
        & \underline{96.7}\,/\,\underline{99.9}\,/\,\textbf{100.0}
        & \textbf{100.0} & \textbf{100.0}
        & 49.2\,/\,\underline{79.2}\,/\,\underline{93.8}
        & \underline{96.7}\,/\,\underline{99.9}\,/\,\textbf{100.0}
        & \textbf{100.0} & \textbf{100.0}
        & 49.2\,/\,\underline{79.2}\,/\,\textbf{93.8}
        & \underline{96.7}\,/\,\underline{99.9}\,/\,\textbf{100.0}
        & \textbf{100.0} & \textbf{100.0}
        & 49.2\,/\,\underline{79.2}\,/\,\underline{93.8}
        & \underline{96.7}\,/\,\underline{99.9}\,/\,\textbf{100.0} \\

        DINOv3~\cite{simeoni2025dinov3}+VGGT~\cite{wang2025vggt}
        & \textbf{100.0} & \textbf{100.0}
        & \underline{49.2}\,/\,\textbf{79.2}\,/\,\underline{93.8}
        & \underline{96.7}\,/\,\underline{99.9}\,/\,\textbf{100.0}
        & \textbf{100.0} & \textbf{100.0}
        & 49.2\,/\,\underline{79.2}\,/\,\underline{93.8}
        & \underline{96.7}\,/\,\underline{99.9}\,/\,\textbf{100.0}
        & \textbf{100.0} & \textbf{100.0}
        & 49.2\,/\,\underline{79.2}\,/\,\textbf{93.8}
        & \underline{96.7}\,/\,\underline{99.9}\,/\,\textbf{100.0}
        & \textbf{100.0} & \textbf{100.0}
        & 49.2\,/\,\underline{79.2}\,/\,\underline{93.8}
        & \underline{96.7}\,/\,\underline{99.9}\,/\,\textbf{100.0} \\

        RobustVGGT~\cite{han2026emergent}
        & \underline{99.3} & \underline{93.7}
        & 49.1\,/\,78.9\,/\,93.2
        & 95.8\,/\,99.3\,/\,99.3
        & \underline{99.6} & 98.9
        & \underline{49.8}\,/\,78.9\,/\,93.3
        & 96.2\,/\,99.4\,/\,99.6
        & \underline{99.4} & 99.1
        & 49.3\,/\,79.0\,/\,\underline{93.2}
        & 95.9\,/\,99.3\,/\,99.4
        & \underline{99.4} & 97.2
        & \underline{49.4}\,/\,78.9\,/\,93.3
        & 96.0\,/\,99.4\,/\,99.4 \\

        RobustPi3~\cite{han2026emergent}
        & 95.0 & \underline{93.0}
        & 10.4\,/\,\underline{52.4}\,/\,\underline{78.4}
        & \textbf{64.7}\,/\,87.2\,/\,93.8
        & 95.6 & \underline{77.9}
        & 8.7\,/\,\underline{43.9}\,/\,\underline{69.3}
        & \underline{55.3}\,/\,81.1\,/\,91.9
        & 96.4 & \underline{76.9}
        & \underline{7.6}\,/\,\underline{41.0}\,/\,\underline{65.2}
        & \underline{53.3}\,/\,\underline{75.8}\,/\,84.0
        & 95.7 & \underline{82.6}
        & 8.9\,/\,\underline{45.8}\,/\,\underline{71.0}
        & \underline{57.8}\,/\,81.4\,/\,89.9 \\

        \midrule
        \rowcolor{gray!20}
        \multicolumn{17}{c}{\textbf{Using Distractor Image as Reference}} \\

        VGGT~\cite{wang2025vggt}
        & \textbf{100.0} & 0.0
        & 0.0\,/\,1.2\,/\,4.7
        & 0.7\,/\,9.1\,/\,24.0
        & \textbf{100.0} & 0.0
        & 0.0\,/\,0.8\,/\,2.4
        & 0.3\,/\,4.2\,/\,13.8
        & \textbf{100.0} & 0.0
        & 0.0\,/\,1.0\,/\,3.0
        & 0.9\,/\,4.8\,/\,14.2
        & \textbf{100.0} & 0.0
        & 0.0\,/\,1.0\,/\,3.4
        & 0.6\,/\,6.0\,/\,17.3 \\

        $\pi^3$~\cite{wang2026pi3}
        & \textbf{100.0} & 0.0
        & \textbf{11.7}\,/\,42.6\,/\,74.6
        & 54.1\,/\,92.4\,/\,\underline{98.2}
        & \textbf{100.0} & 0.0
        & 8.9\,/\,33.0\,/\,61.2
        & 46.1\,/\,85.8\,/\,\underline{96.8}
        & \textbf{100.0} & 0.0
        & 6.3\,/\,28.1\,/\,54.3
        & 35.6\,/\,73.4\,/\,\underline{85.3}
        & \textbf{100.0} & 0.0
        & \underline{9.0}\,/\,34.6\,/\,63.4
        & 45.3\,/\,83.9\,/\,\underline{93.4} \\

        MegaLoc~\cite{Berton_2025_MegaLoc}+VGGT~\cite{wang2025vggt}
        & 0.1 & 79.0
        & 0.0\,/\,0.0\,/\,0.0
        & 0.0\,/\,0.0\,/\,0.0
        & 0.1 & 79.6
        & 0.0\,/\,0.0\,/\,0.0
        & 0.0\,/\,0.0\,/\,0.0
        & 0.1 & 79.7
        & 0.0\,/\,0.0\,/\,0.0
        & 0.0\,/\,0.0\,/\,0.0
        & 0.1 & 79.4
        & 0.0\,/\,0.0\,/\,0.0
        & 0.0\,/\,0.0\,/\,0.0 \\

        DINOv2$^\dagger$\cite{oquab2024dinov2}+VGGT~\cite{wang2025vggt}
        & 0.4 & 80.0
        & 0.0\,/\,0.0\,/\,0.0
        & 0.0\,/\,0.0\,/\,0.2
        & 0.4 & 80.1
        & 0.0\,/\,0.0\,/\,0.0
        & 0.0\,/\,0.0\,/\,0.0
        & 0.4 & 80.1
        & 0.0\,/\,0.0\,/\,0.0
        & 0.0\,/\,0.0\,/\,0.0
        & 0.4 & 80.1
        & 0.0\,/\,0.0\,/\,0.0
        & 0.0\,/\,0.0\,/\,0.1 \\

        DINOv3~\cite{simeoni2025dinov3}+VGGT~\cite{wang2025vggt}
        & 0.0 & 80.0
        & 0.0\,/\,0.0\,/\,0.0
        & 0.0\,/\,0.0\,/\,0.0
        & 0.0 & 80.0
        & 0.0\,/\,0.0\,/\,0.0
        & 0.0\,/\,0.0\,/\,0.0
        & 0.0 & 80.0
        & 0.0\,/\,0.0\,/\,0.0
        & 0.0\,/\,0.0\,/\,0.0
        & 0.0 & 80.0
        & 0.0\,/\,0.0\,/\,0.0
        & 0.0\,/\,0.0\,/\,0.0 \\

        RobustVGGT~\cite{han2026emergent}
        & 0.2 & 78.3
        & 0.0\,/\,0.0\,/\,0.0
        & 0.0\,/\,0.0\,/\,0.0
        & 0.0 & 78.9
        & 0.0\,/\,0.0\,/\,0.0
        & 0.0\,/\,0.0\,/\,0.0
        & 0.1 & 78.9
        & 0.0\,/\,0.0\,/\,0.0
        & 0.0\,/\,0.0\,/\,0.0
        & 0.1 & 78.7
        & 0.0\,/\,0.0\,/\,0.0
        & 0.0\,/\,0.0\,/\,0.0 \\

        RobustPi3~\cite{han2026emergent}
        & 36.3 & 35.3
        & 3.4\,/\,20.1\,/\,30.9
        & 23.0\,/\,32.1\,/\,34.9
        & 35.4 & 40.7
        & 3.0\,/\,14.3\,/\,27.0
        & 17.2\,/\,26.0\,/\,30.6
        & 39.2 & 42.2
        & 4.0\,/\,16.1\,/\,27.3
        & 19.9\,/\,29.2\,/\,33.2
        & 37.0 & 39.4
        & 3.5\,/\,16.9\,/\,28.4
        & 20.0\,/\,29.1\,/\,32.9 \\

        \midrule
        \rowcolor{gray!20}
        \multicolumn{17}{c}{\textbf{Automatic Reference Selection}} \\

        \textbf{VGGT~\cite{wang2025vggt} + FreeView (Ours)}
        & \textbf{100.0} & \textbf{100.0}
        & \textbf{53.2}\,/\,\underline{79.0}\,/\,\textbf{94.3}
        & \textbf{97.3}\,/\,\textbf{100.0}\,/\,\textbf{100.0}
        & \textbf{100.0} & \textbf{100.0}
        & \textbf{55.2}\,/\,\textbf{81.3}\,/\,\textbf{94.0}
        & \textbf{97.3}\,/\,\textbf{100.0}\,/\,\textbf{100.0}
        & \textbf{100.0} & 99.7
        & \textbf{53.2}\,/\,\textbf{82.0}\,/\,\textbf{93.8}
        & \textbf{97.3}\,/\,\textbf{100.0}\,/\,\textbf{100.0}
        & \textbf{100.0} & \underline{99.9}
        & \textbf{53.9}\,/\,\textbf{80.8}\,/\,\textbf{94.0}
        & \textbf{97.3}\,/\,\textbf{100.0}\,/\,\textbf{100.0} \\

        \textbf{$\pi^3$~\cite{wang2026pi3} + FreeView (Ours)}
        & \underline{99.4} & \textbf{99.3}
        & \underline{11.6}\,/\,\textbf{57.8}\,/\,\textbf{81.9}
        & \underline{61.6}\,/\,\textbf{95.0}\,/\,\textbf{98.3}
        & \underline{99.4} & \textbf{97.3}
        & \textbf{11.9}\,/\,\textbf{57.4}\,/\,\textbf{81.2}
        & \textbf{62.2}\,/\,\textbf{95.1}\,/\,\textbf{98.3}
        & \underline{99.4} & \textbf{96.2}
        & \textbf{12.6}\,/\,\textbf{55.3}\,/\,\textbf{81.1}
        & \textbf{63.6}\,/\,\textbf{95.1}\,/\,\textbf{98.3}
        & \underline{99.4} & \textbf{97.6}
        & \textbf{12.0}\,/\,\textbf{56.9}\,/\,\textbf{81.4}
        & \textbf{62.4}\,/\,\textbf{95.1}\,/\,\textbf{98.3} \\

        \bottomrule
    \end{tabular}}
    \end{subtable}
    \par\medskip

    \begin{subtable}{\textwidth}
        \centering
        \caption{ETH3D}
        \label{tab:pose_eval_eth3d}
        \resizebox{\linewidth}{!}{%
\begin{tabular}{l|cccc|cccc|cccc|cccc}
        \toprule
        \multirow{2}{*}{Methods}
        & \multicolumn{4}{c|}{Small} 
        & \multicolumn{4}{c|}{Medium} 
        & \multicolumn{4}{c|}{Large} 
        & \multicolumn{4}{c}{Avg} \\

        & CK$\uparrow$
        & DR$\uparrow$
        & T@2\,/\,5\,/\,10\%$\uparrow$
        & R@2\,/\,5\,/\,10$^\circ$$\uparrow$
        & CK$\uparrow$
        & DR$\uparrow$
        & T@2\,/\,5\,/\,10\%$\uparrow$
        & R@2\,/\,5\,/\,10$^\circ$$\uparrow$
        & CK$\uparrow$
        & DR$\uparrow$
        & T@2\,/\,5\,/\,10\%$\uparrow$
        & R@2\,/\,5\,/\,10$^\circ$$\uparrow$
        & CK$\uparrow$
        & DR$\uparrow$
        & T@2\,/\,5\,/\,10\%$\uparrow$
        & R@2\,/\,5\,/\,10$^\circ$$\uparrow$ \\
        \midrule
\rowcolor{gray!20}
\multicolumn{17}{c}{\textbf{Oracle (Using Clean Images Only; No Distractors)}} \\
\rowcolor{gray!10}
VGGT (Clean-only)~\cite{wang2025vggt}
 & 100.0 & -- & 11.9\,/\,35.7\,/\,56.8 & 49.6\,/\,68.2\,/\,76.8
 & 100.0 & -- & 11.9\,/\,35.7\,/\,56.8 & 49.6\,/\,68.2\,/\,76.8
 & 100.0 & -- & 11.9\,/\,35.7\,/\,56.8 & 49.6\,/\,68.2\,/\,76.8
 & 100.0 & -- & 11.9\,/\,35.7\,/\,56.8 & 49.6\,/\,68.2\,/\,76.8 \\
\rowcolor{gray!10}
$\pi^3$ (Clean-only)~\cite{wang2026pi3}
 & 100.0 & -- & 28.5\,/\,53.6\,/\,66.9 & 61.1\,/\,91.7\,/\,98.8
 & 100.0 & -- & 28.5\,/\,53.6\,/\,66.9 & 61.1\,/\,91.7\,/\,98.8
 & 100.0 & -- & 28.5\,/\,53.6\,/\,66.9 & 61.1\,/\,91.7\,/\,98.8
 & 100.0 & -- & 28.5\,/\,53.6\,/\,66.9 & 61.1\,/\,91.7\,/\,98.8 \\
\midrule
\rowcolor{gray!20}
\multicolumn{17}{c}{\textbf{Using Clean Image as Reference}} \\
VGGT~\cite{wang2025vggt}
 & \textbf{100.0} & 0.0 & 6.5\,/\,28.9\,/\,53.7 & 40.8\,/\,67.0\,/\,\textbf{79.4}
 & \textbf{100.0} & 0.0 & 3.1\,/\,19.5\,/\,42.0 & 29.9\,/\,54.8\,/\,69.3
 & \textbf{100.0} & 0.0 & 1.1\,/\,14.0\,/\,32.1 & 21.2\,/\,45.0\,/\,60.0
 & \textbf{100.0} & 0.0 & 3.6\,/\,20.8\,/\,42.6 & 30.6\,/\,55.6\,/\,69.6 \\
$\pi^3$~\cite{wang2026pi3}
 & \textbf{100.0} & 0.0 & 12.4\,/\,43.5\,/\,\underline{61.0} & 46.2\,/\,\textbf{78.6}\,/\,\underline{90.4}
 & \textbf{100.0} & 0.0 & 3.6\,/\,18.9\,/\,40.0 & 29.0\,/\,53.0\,/\,74.4
 & \textbf{100.0} & 0.0 & 2.1\,/\,8.0\,/\,31.4 & 19.5\,/\,32.4\,/\,53.0
 & \textbf{100.0} & 0.0 & 6.0\,/\,23.5\,/\,44.1 & 31.6\,/\,54.6\,/\,72.6 \\
MegaLoc~\cite{Berton_2025_MegaLoc}+VGGT~\cite{wang2025vggt}
 & 85.7 & 95.0 & 14.0\,/\,33.9\,/\,50.0 & 40.5\,/\,59.5\,/\,65.5
 & 85.7 & 93.5 & \textbf{15.0}\,/\,33.6\,/\,49.2 & 39.3\,/\,57.5\,/\,64.9
 & 85.7 & 93.8 & \underline{14.4}\,/\,31.1\,/\,46.8 & 36.5\,/\,56.8\,/\,63.6
 & 85.7 & 94.1 & 14.5\,/\,32.9\,/\,48.7 & 38.8\,/\,57.9\,/\,64.6 \\
DINOv2$^\dagger$\cite{oquab2024dinov2}+VGGT~\cite{wang2025vggt}
 & 80.6 & 95.7 & \underline{15.5}\,/\,38.3\,/\,54.6 & 48.2\,/\,58.1\,/\,59.6
 & 80.6 & 93.6 & \underline{14.9}\,/\,\textbf{37.7}\,/\,53.6 & 45.4\,/\,58.5\,/\,60.8
 & 80.6 & 93.2 & \textbf{14.8}\,/\,\underline{36.5}\,/\,50.8 & 43.3\,/\,54.4\,/\,59.3
 & 80.6 & 94.1 & \underline{15.0}\,/\,\underline{37.5}\,/\,53.0 & 45.6\,/\,57.0\,/\,59.9 \\
DINOv3~\cite{simeoni2025dinov3}+VGGT~\cite{wang2025vggt}
 & 80.7 & 96.7 & 13.1\,/\,36.8\,/\,53.0 & 47.0\,/\,60.0\,/\,62.5
 & 80.7 & 95.1 & 13.2\,/\,36.0\,/\,52.4 & \underline{46.7}\,/\,58.5\,/\,62.9
 & 80.7 & 93.9 & 12.1\,/\,33.0\,/\,48.6 & 42.3\,/\,53.1\,/\,61.0
 & 80.7 & 95.2 & 12.8\,/\,35.2\,/\,51.3 & 45.3\,/\,57.2\,/\,62.1 \\
RobustVGGT~\cite{han2026emergent}
 & 88.3 & \underline{99.7} & \underline{15.5}\,/\,\textbf{41.4}\,/\,\textbf{60.4} & \underline{50.2}\,/\,\textbf{70.0}\,/\,\underline{76.7}
 & 94.2 & \textbf{99.4} & 12.5\,/\,\underline{37.5}\,/\,\textbf{58.5} & 46.2\,/\,\textbf{71.1}\,/\,\textbf{76.5}
 & 96.8 & \textbf{99.6} & 12.5\,/\,\textbf{37.4}\,/\,\underline{56.5} & \underline{46.9}\,/\,\underline{68.1}\,/\,\textbf{76.0}
 & 93.1 & \textbf{99.5} & 13.5\,/\,\textbf{38.8}\,/\,\textbf{58.5} & \underline{47.8}\,/\,\textbf{69.7}\,/\,\textbf{76.4} \\
RobustPi3~\cite{han2026emergent}
 & \underline{93.2} & \underline{95.7} & \underline{27.7}\,/\,\underline{48.6}\,/\,60.2 & \underline{57.0}\,/\,\underline{78.2}\,/\,83.7
 & \underline{97.1} & \underline{88.3} & \underline{27.4}\,/\,\underline{48.7}\,/\,\underline{58.3} & \textbf{61.4}\,/\,\textbf{82.0}\,/\,\textbf{87.9}
 & \underline{98.3} & \underline{73.9} & \textbf{18.5}\,/\,\underline{41.5}\,/\,\textbf{54.4} & \underline{46.1}\,/\,\textbf{66.0}\,/\,\textbf{81.7}
 & \underline{96.2} & \underline{86.0} & \underline{24.5}\,/\,\underline{46.3}\,/\,\underline{57.7} & \underline{54.8}\,/\,\textbf{75.4}\,/\,\textbf{84.4} \\
\midrule
\rowcolor{gray!20}
\multicolumn{17}{c}{\textbf{Using Distractor Image as Reference}} \\
VGGT~\cite{wang2025vggt}
 & \textbf{100.0} & 0.0 & 0.1\,/\,0.8\,/\,3.1 & 0.7\,/\,5.5\,/\,8.0
 & \textbf{100.0} & 0.0 & 0.1\,/\,0.8\,/\,2.1 & 0.1\,/\,3.1\,/\,5.7
 & \textbf{100.0} & 0.0 & 0.1\,/\,0.4\,/\,1.7 & 0.0\,/\,1.1\,/\,4.4
 & \textbf{100.0} & 0.0 & 0.1\,/\,0.7\,/\,2.3 & 0.3\,/\,3.2\,/\,6.0 \\
$\pi^3$~\cite{wang2026pi3}
 & \textbf{100.0} & 0.0 & 12.6\,/\,43.5\,/\,\underline{61.0} & 46.2\,/\,\textbf{78.6}\,/\,\textbf{90.5}
 & \textbf{100.0} & 0.0 & 3.6\,/\,19.0\,/\,40.5 & 29.2\,/\,53.0\,/\,74.4
 & \textbf{100.0} & 0.0 & 2.1\,/\,8.0\,/\,31.5 & 19.6\,/\,32.1\,/\,53.1
 & \textbf{100.0} & 0.0 & 6.1\,/\,23.5\,/\,44.3 & 31.7\,/\,54.6\,/\,72.7 \\
MegaLoc~\cite{Berton_2025_MegaLoc}+VGGT~\cite{wang2025vggt}
 & 5.4 & 77.3 & 0.0\,/\,0.1\,/\,0.4 & 0.0\,/\,0.0\,/\,0.1
 & 5.4 & 84.8 & 0.0\,/\,0.0\,/\,0.0 & 0.0\,/\,0.0\,/\,0.2
 & 5.4 & 86.9 & 0.0\,/\,0.1\,/\,0.4 & 0.0\,/\,0.0\,/\,0.2
 & 5.4 & 83.0 & 0.0\,/\,0.1\,/\,0.2 & 0.0\,/\,0.0\,/\,0.2 \\
DINOv2$^\dagger$\cite{oquab2024dinov2}+VGGT~\cite{wang2025vggt}
 & 5.2 & 75.3 & 0.0\,/\,0.1\,/\,0.2 & 0.0\,/\,0.0\,/\,0.2
 & 5.2 & 82.5 & 0.0\,/\,0.2\,/\,0.2 & 0.0\,/\,0.0\,/\,0.0
 & 5.2 & 85.1 & 0.0\,/\,0.1\,/\,0.2 & 0.0\,/\,0.4\,/\,0.4
 & 5.2 & 81.0 & 0.0\,/\,0.2\,/\,0.2 & 0.0\,/\,0.1\,/\,0.2 \\
DINOv3~\cite{simeoni2025dinov3}+VGGT~\cite{wang2025vggt}
 & 4.2 & 74.7 & 0.1\,/\,0.1\,/\,0.5 & 0.0\,/\,0.0\,/\,0.0
 & 4.2 & 83.2 & 0.1\,/\,0.1\,/\,0.4 & 0.0\,/\,0.0\,/\,0.0
 & 4.2 & 85.7 & 0.0\,/\,0.0\,/\,0.4 & 0.0\,/\,0.0\,/\,0.0
 & 4.2 & 81.2 & 0.1\,/\,0.1\,/\,0.4 & 0.0\,/\,0.0\,/\,0.0 \\
RobustVGGT~\cite{han2026emergent}
 & 0.7 & 72.0 & 0.0\,/\,0.0\,/\,0.0 & 0.0\,/\,0.0\,/\,0.0
 & 1.1 & 82.1 & 0.0\,/\,0.0\,/\,0.0 & 0.0\,/\,0.0\,/\,0.0
 & 0.7 & 86.4 & 0.4\,/\,0.4\,/\,0.4 & 0.0\,/\,0.0\,/\,0.0
 & 0.8 & 80.2 & 0.1\,/\,0.1\,/\,0.1 & 0.0\,/\,0.0\,/\,0.0 \\
RobustPi3~\cite{han2026emergent}
 & 13.0 & 27.3 & 1.8\,/\,3.0\,/\,4.8 & 4.3\,/\,6.2\,/\,8.1
 & 26.1 & 36.5 & 2.5\,/\,6.5\,/\,9.4 & 10.6\,/\,15.2\,/\,18.5
 & 36.8 & 46.9 & 1.4\,/\,4.5\,/\,10.0 & 6.4\,/\,13.2\,/\,19.4
 & 25.3 & 36.9 & 1.9\,/\,4.7\,/\,8.1 & 7.1\,/\,11.5\,/\,15.3 \\
\midrule
\rowcolor{gray!20}
\multicolumn{17}{c}{\textbf{Automatic Reference Selection}} \\
\textbf{VGGT~\cite{wang2025vggt} + FreeView (Ours)}
 & \underline{93.5} & \textbf{100.0} & \textbf{17.6}\,/\,\underline{41.2}\,/\,\underline{55.4} & \textbf{56.9}\,/\,\underline{67.3}\,/\,68.5
 & \underline{96.4} & \underline{96.4} & 13.2\,/\,35.8\,/\,\underline{55.4} & \textbf{50.0}\,/\,\underline{67.9}\,/\,\underline{70.8}
 & \underline{99.4} & \underline{95.8} & \underline{14.4}\,/\,34.9\,/\,\textbf{57.1} & \textbf{51.3}\,/\,\textbf{70.8}\,/\,\underline{73.8}
 & \underline{96.4} & \underline{97.4} & \textbf{15.1}\,/\,37.3\,/\,\underline{56.0} & \textbf{52.7}\,/\,\underline{68.7}\,/\,\underline{71.0} \\
\textbf{$\pi^3$~\cite{wang2026pi3} + FreeView (Ours)}
 & 82.7 & \textbf{96.7} & \textbf{31.5}\,/\,\textbf{56.5}\,/\,\textbf{65.0} & \textbf{60.4}\,/\,66.1\,/\,73.2
 & 91.1 & \textbf{91.7} & \textbf{32.1}\,/\,\textbf{56.0}\,/\,\textbf{63.7} & \underline{60.8}\,/\,\underline{70.0}\,/\,\underline{81.0}
 & 93.5 & \textbf{84.4} & \underline{18.0}\,/\,\textbf{45.1}\,/\,\underline{54.2} & \textbf{47.0}\,/\,\underline{65.2}\,/\,\underline{81.0}
 & 89.1 & \textbf{90.9} & \textbf{27.2}\,/\,\textbf{52.5}\,/\,\textbf{61.0} & \textbf{56.1}\,/\,\underline{67.1}\,/\,\underline{78.4} \\
\bottomrule
\end{tabular}}
    \end{subtable}
    \par\medskip

    \begin{subtable}{\textwidth}
        \centering
        \caption{On-the-Go}
        \label{tab:pose_eval_otg}
        \resizebox{\linewidth}{!}{%
\begin{tabular}{l|cccc|cccc|cccc|cccc}
        \toprule
        \multirow{2}{*}{Methods}
        & \multicolumn{4}{c|}{Small} 
        & \multicolumn{4}{c|}{Medium} 
        & \multicolumn{4}{c|}{Large} 
        & \multicolumn{4}{c}{Avg} \\

        & CK$\uparrow$
        & DR$\uparrow$
        & T@2\,/\,5\,/\,10\%$\uparrow$
        & R@2\,/\,5\,/\,10$^\circ$$\uparrow$
        & CK$\uparrow$
        & DR$\uparrow$
        & T@2\,/\,5\,/\,10\%$\uparrow$
        & R@2\,/\,5\,/\,10$^\circ$$\uparrow$
        & CK$\uparrow$
        & DR$\uparrow$
        & T@2\,/\,5\,/\,10\%$\uparrow$
        & R@2\,/\,5\,/\,10$^\circ$$\uparrow$
        & CK$\uparrow$
        & DR$\uparrow$
        & T@2\,/\,5\,/\,10\%$\uparrow$
        & R@2\,/\,5\,/\,10$^\circ$$\uparrow$ \\
        \midrule
\rowcolor{gray!20}
\multicolumn{17}{c}{\textbf{Oracle (Using Clean Images Only; No Distractors)}} \\

        \rowcolor{gray!10}
        VGGT (Clean-only)~\cite{wang2025vggt}
        & 100.0 & -- & 74.9\,/\,95.8\,/\,99.8 & 92.9\,/\,99.9\,/\,100.0
        & 100.0 & -- & 74.9\,/\,95.8\,/\,99.8 & 92.9\,/\,99.9\,/\,100.0
        & 100.0 & -- & 74.9\,/\,95.8\,/\,99.8 & 92.9\,/\,99.9\,/\,100.0
        & 100.0 & -- & 74.9\,/\,95.8\,/\,99.8 & 92.9\,/\,99.9\,/\,100.0 \\

        \rowcolor{gray!10}
        $\pi^3$ (Clean-only)~\cite{wang2026pi3}
        & 100.0 & -- & 70.8\,/\,95.8\,/\,99.2 & 87.8\,/\,100.0\,/\,100.0
        & 100.0 & -- & 70.8\,/\,95.8\,/\,99.2 & 87.8\,/\,100.0\,/\,100.0
        & 100.0 & -- & 70.8\,/\,95.8\,/\,99.2 & 87.8\,/\,100.0\,/\,100.0
        & 100.0 & -- & 70.8\,/\,95.8\,/\,99.2 & 87.8\,/\,100.0\,/\,100.0 \\

        \rowcolor{gray!20}
        \multicolumn{17}{c}{\textbf{Using Clean Image as Reference}} \\

        VGGT~\cite{wang2025vggt}
        & \textbf{100.0} & 0.0
        & 59.0\,/\,91.3\,/\,97.2
        & 89.9\,/\,97.4\,/\,97.5
        & \textbf{100.0} & 0.0
        & 41.6\,/\,85.2\,/\,95.1
        & 84.3\,/\,97.1\,/\,97.5
        & \textbf{100.0} & 0.0
        & 34.5\,/\,76.3\,/\,90.3
        & 76.3\,/\,88.8\,/\,\underline{99.3}
        & \textbf{100.0} & 0.0
        & 45.0\,/\,84.3\,/\,94.2
        & 83.5\,/\,94.4\,/\,98.1 \\

        $\pi^3$~\cite{wang2026pi3}
        & \textbf{100.0} & 0.0
        & 50.4\,/\,87.9\,/\,\underline{97.3}
        & 76.7\,/\,\textbf{100.0}\,/\,\textbf{100.0}
        & \textbf{100.0} & 0.0
        & 25.2\,/\,74.0\,/\,88.3
        & 71.0\,/\,84.9\,/\,90.5
        & \textbf{100.0} & 0.0
        & 12.3\,/\,49.5\,/\,72.2
        & 53.5\,/\,66.9\,/\,87.2
        & \textbf{100.0} & 0.0
        & 29.3\,/\,70.5\,/\,85.9
        & 67.1\,/\,83.9\,/\,92.6 \\

        MegaLoc~\cite{Berton_2025_MegaLoc}+VGGT~\cite{wang2025vggt}
        & 95.6 & 97.3
        & 71.5\,/\,91.4\,/\,95.3
        & 87.3\,/\,95.5\,/\,95.6
        & 95.6 & 97.3
        & 71.2\,/\,91.5\,/\,95.4
        & 88.4\,/\,95.5\,/\,95.6
        & 95.6 & 97.6
        & \underline{71.2}\,/\,\underline{91.5}\,/\,\underline{95.3}
        & \textbf{89.1}\,/\,95.5\,/\,95.6
        & 95.6 & 97.4
        & 71.3\,/\,91.5\,/\,95.3
        & \underline{88.3}\,/\,95.5\,/\,95.6 \\

        DINOv2$^\dagger$\cite{oquab2024dinov2}+VGGT~\cite{wang2025vggt}
        & 94.8 & 98.0
        & 71.3\,/\,90.9\,/\,94.5
        & 87.9\,/\,94.8\,/\,94.8
        & 94.8 & 97.8
        & 70.5\,/\,91.0\,/\,94.5
        & 88.3\,/\,94.8\,/\,94.8
        & 94.8 & \underline{98.1}
        & 70.2\,/\,89.6\,/\,93.6
        & \underline{86.8}\,/\,94.2\,/\,94.8
        & 94.8 & 97.9
        & 70.7\,/\,90.5\,/\,94.2
        & 87.7\,/\,94.6\,/\,94.8 \\

        DINOv3~\cite{simeoni2025dinov3}+VGGT~\cite{wang2025vggt}
        & 98.4 & 91.5
        & 74.0\,/\,94.1\,/\,98.3
        & \textbf{91.4}\,/\,98.3\,/\,98.4
        & 98.4 & 92.6
        & \underline{72.0}\,/\,94.1\,/\,\underline{98.3}
        & \underline{91.4}\,/\,98.3\,/\,98.4
        & 98.4 & 93.3
        & 68.5\,/\,87.3\,/\,93.8
        & 81.8\,/\,95.6\,/\,98.4
        & 98.4 & 92.5
        & 71.5\,/\,91.8\,/\,96.8
        & 88.2\,/\,97.4\,/\,98.4 \\

        RobustVGGT~\cite{han2026emergent}
        & \underline{99.4} & \textbf{100.0}
        & \underline{74.3}\,/\,\underline{95.3}\,/\,\textbf{99.2}
        & \underline{91.3}\,/\,\underline{99.4}\,/\,\underline{99.4}
        & \underline{99.7} & \underline{98.0}
        & 71.2\,/\,\underline{94.5}\,/\,\textbf{99.2}
        & 90.3\,/\,\textbf{99.7}\,/\,\textbf{99.7}
        & \underline{99.9} & 96.7
        & 69.8\,/\,88.0\,/\,94.8
        & 82.3\,/\,\underline{97.2}\,/\,\textbf{99.9}
        & \underline{99.7} & \underline{98.2}
        & \underline{71.8}\,/\,\underline{92.6}\,/\,\underline{97.7}
        & 88.0\,/\,\underline{98.8}\,/\,\textbf{99.7} \\

        RobustPi3~\cite{han2026emergent}
        & \underline{99.5} & \underline{93.3}
        & \textbf{68.7}\,/\,\textbf{94.2}\,/\,\textbf{98.8}
        & \textbf{86.9}\,/\,\underline{99.5}\,/\,\underline{99.5}
        & \underline{99.4} & \underline{87.0}
        & \textbf{66.6}\,/\,\textbf{90.5}\,/\,\textbf{98.6}
        & \textbf{84.0}\,/\,\textbf{99.4}\,/\,\textbf{99.4}
        & 95.8 & \underline{80.4}
        & \underline{62.1}\,/\,\underline{82.9}\,/\,\underline{91.9}
        & \underline{74.8}\,/\,\underline{93.7}\,/\,\underline{94.1}
        & \underline{98.2} & \underline{86.9}
        & \underline{65.8}\,/\,\underline{89.2}\,/\,\textbf{96.4}
        & \underline{81.9}\,/\,\textbf{97.5}\,/\,\textbf{97.7} \\

        \midrule
        \rowcolor{gray!20}
        \multicolumn{17}{c}{\textbf{Using Distractor Image as Reference}} \\

        VGGT~\cite{wang2025vggt}
        & \textbf{100.0} & 0.0
        & 0.1\,/\,1.2\,/\,4.3
        & 0.8\,/\,8.3\,/\,18.9
        & \textbf{100.0} & 0.0
        & 0.0\,/\,0.4\,/\,1.8
        & 0.2\,/\,1.8\,/\,6.5
        & \textbf{100.0} & 0.0
        & 0.0\,/\,0.2\,/\,1.7
        & 0.3\,/\,2.1\,/\,5.1
        & \textbf{100.0} & 0.0
        & 0.0\,/\,0.6\,/\,2.6
        & 0.4\,/\,4.1\,/\,10.2 \\

        $\pi^3$~\cite{wang2026pi3}
        & \textbf{100.0} & 0.0
        & 50.3\,/\,88.1\,/\,97.2
        & 76.8\,/\,\textbf{100.0}\,/\,\textbf{100.0}
        & \textbf{100.0} & 0.0
        & 24.7\,/\,74.3\,/\,88.3
        & 71.2\,/\,84.9\,/\,90.5
        & \textbf{100.0} & 0.0
        & 12.4\,/\,49.3\,/\,72.5
        & 53.4\,/\,66.8\,/\,87.2
        & \textbf{100.0} & 0.0
        & 29.1\,/\,70.6\,/\,86.0
        & 67.1\,/\,83.9\,/\,92.6 \\

        MegaLoc~\cite{Berton_2025_MegaLoc}+VGGT~\cite{wang2025vggt}
        & 2.4 & 83.3
        & 0.0\,/\,0.0\,/\,0.3
        & 0.0\,/\,0.3\,/\,1.1
        & 2.4 & 85.2
        & 0.0\,/\,0.1\,/\,0.5
        & 0.3\,/\,0.6\,/\,1.1
        & 2.4 & 85.3
        & 0.1\,/\,0.1\,/\,0.8
        & 0.2\,/\,0.8\,/\,1.3
        & 2.4 & 84.6
        & 0.0\,/\,0.1\,/\,0.5
        & 0.2\,/\,0.6\,/\,1.2 \\

        DINOv2$^\dagger$\cite{oquab2024dinov2}+VGGT~\cite{wang2025vggt}
        & 1.8 & 84.0
        & 0.0\,/\,0.1\,/\,0.3
        & 0.0\,/\,0.1\,/\,0.3
        & 1.8 & 85.1
        & 0.0\,/\,0.3\,/\,1.3
        & 0.3\,/\,1.0\,/\,1.3
        & 1.8 & 85.3
        & 0.0\,/\,0.2\,/\,0.8
        & 0.4\,/\,0.8\,/\,1.3
        & 1.8 & 84.8
        & 0.0\,/\,0.2\,/\,0.8
        & 0.3\,/\,0.6\,/\,0.9 \\

        DINOv3~\cite{simeoni2025dinov3}+VGGT~\cite{wang2025vggt}
        & 6.4 & 78.3
        & 0.0\,/\,0.1\,/\,2.1
        & 1.4\,/\,3.5\,/\,4.8
        & 6.4 & 78.2
        & 0.1\,/\,0.7\,/\,1.8
        & 0.3\,/\,2.5\,/\,3.3
        & 6.4 & 78.6
        & 0.1\,/\,0.8\,/\,2.3
        & 1.3\,/\,3.0\,/\,3.7
        & 6.4 & 78.3
        & 0.1\,/\,0.5\,/\,2.0
        & 1.0\,/\,3.0\,/\,3.9 \\

        RobustVGGT~\cite{han2026emergent}
        & 0.0 & 79.5
        & 0.0\,/\,0.0\,/\,0.0
        & 0.0\,/\,0.0\,/\,0.0
        & 0.1 & 81.8
        & 0.0\,/\,0.0\,/\,0.0
        & 0.0\,/\,0.0\,/\,0.0
        & 0.2 & 82.4
        & 0.0\,/\,0.0\,/\,0.0
        & 0.0\,/\,0.0\,/\,0.0
        & 0.1 & 81.2
        & 0.0\,/\,0.0\,/\,0.0
        & 0.0\,/\,0.0\,/\,0.0 \\

        RobustPi3~\cite{han2026emergent}
        & 14.0 & 29.5
        & 9.9\,/\,13.3\,/\,13.6
        & 12.7\,/\,13.3\,/\,13.3
        & 17.5 & 46.0
        & 8.1\,/\,12.4\,/\,14.7
        & 12.5\,/\,13.9\,/\,14.3
        & 26.0 & 51.5
        & 12.6\,/\,20.2\,/\,22.7
        & 18.1\,/\,22.6\,/\,23.1
        & 19.2 & 42.3
        & 10.2\,/\,15.3\,/\,17.0
        & 14.4\,/\,16.6\,/\,16.9 \\

        \midrule
        \rowcolor{gray!20}
        \multicolumn{17}{c}{\textbf{Automatic Reference Selection}} \\

        \textbf{VGGT~\cite{wang2025vggt} + FreeView (Ours)}
        & \textbf{100.0} & \underline{99.3}
        & \textbf{79.7}\,/\,\textbf{95.9}\,/\,\underline{98.9}
        & 87.5\,/\,\textbf{100.0}\,/\,\textbf{100.0}
        & 98.8 & \textbf{99.6}
        & \textbf{77.4}\,/\,\textbf{95.2}\,/\,97.9
        & \textbf{92.9}\,/\,\underline{98.8}\,/\,\underline{98.8}
        & 98.8 & \textbf{98.5}
        & \textbf{73.9}\,/\,\textbf{94.4}\,/\,\textbf{97.5}
        & 86.6\,/\,\textbf{98.8}\,/\,98.8
        & 99.2 & \textbf{99.1}
        & \textbf{77.0}\,/\,\textbf{95.2}\,/\,\textbf{98.1}
        & \textbf{89.0}\,/\,\textbf{99.2}\,/\,\underline{99.2} \\

        \textbf{$\pi^3$~\cite{wang2026pi3} + FreeView (Ours)}
        & 95.4 & \textbf{100.0}
        & \underline{68.3}\,/\,\underline{91.1}\,/\,94.6
        & \underline{83.3}\,/\,95.4\,/\,95.4
        & 95.8 & \textbf{92.5}
        & \underline{64.6}\,/\,\underline{89.3}\,/\,\underline{94.6}
        & \underline{82.3}\,/\,\underline{95.8}\,/\,\underline{95.8}
        & \underline{99.2} & \textbf{90.8}
        & \textbf{66.8}\,/\,\textbf{90.4}\,/\,\textbf{98.3}
        & \textbf{83.7}\,/\,\textbf{99.2}\,/\,\textbf{99.2}
        & 96.8 & \textbf{94.4}
        & \textbf{66.6}\,/\,\textbf{90.3}\,/\,\underline{95.8}
        & \textbf{83.1}\,/\,\underline{96.8}\,/\,\underline{96.8} \\

        \bottomrule
    \end{tabular}}
    \end{subtable}

\end{table*}

%% file: ablation/module.tex
\begin{table}[t]
    \centering
    \caption{Ablation study on Consensus Reference Selection (CRS) and Adaptive Dual-Cue Rejection (ADCR).}
    \vspace{-5pt}
    \label{tab:component_ablation}

    \resizebox{\columnwidth}{!}{
    \begin{tabular}{cc|cccc}
        \toprule
        
        CRS
        & ADCR
        & CK$\uparrow$
        & DR$\uparrow$
        & T@2/5/10\%$\uparrow$
        & R@2/5/10$^\circ\uparrow$ \\
        \midrule

        \rowcolor{gray!20}
        \multicolumn{6}{c}{\textbf{Phototourism}} \\

     \xmark
        & \xmark
        & 0.1
        & 78.9
        & 0.0\,/\,0.0\,/\,0.0
        & 0.0\,/\,0.0\,/\,0.0 \\

      \cmark
        & \xmark
        & \textbf{100.0}
        & 99.0
        & 52.8\,/\,81.7\,/\,93.3
        & 97.2\,/\,\textbf{100.0}\,/\,\textbf{100.0} \\

        \rowcolor{gray!10}
 \cmark
        & \cmark
        & \textbf{100.0}
        & \textbf{99.7}
        & \textbf{53.2}\,/\,\textbf{82.0}\,/\,\textbf{93.8}
        & \textbf{97.3}\,/\,\textbf{100.0}\,/\,\textbf{100.0} \\

        \midrule

        \rowcolor{gray!20}
        \multicolumn{6}{c}{\textbf{ETH3D}} \\

 \xmark
        & \xmark
        & 0.7
        & 86.4
        & 0.4\,/\,0.4\,/\,0.4
        & 0.0\,/\,0.0\,/\,0.0 \\
 \cmark
        & \xmark
        & 98.8
        & \textbf{98.9}
        & 12.5\,/\,\textbf{38.7}\,/\,\textbf{63.1}
        & \textbf{54.2}\,/\,\textbf{71.4}\,/\,\textbf{73.8} \\

        \rowcolor{gray!10}
 \cmark
        & \cmark
        & \textbf{99.4}
        & 95.8
        & \textbf{14.4}\,/\,34.9\,/\,57.1
        & 51.3\,/\,70.8\,/\,\textbf{73.8} \\

        \midrule

        \rowcolor{gray!20}
        \multicolumn{6}{c}{\textbf{On-the-Go}} \\

  \xmark
        & \xmark
        & 0.2
        & 82.4
        & 0.0\,/\,0.0\,/\,0.0
        & 0.0\,/\,0.0\,/\,0.0 \\

    \cmark
        & \xmark
        & \textbf{100.0}
        & 96.3
        & 69.2\,/\,86.3\,/\,89.6
        & 79.6\,/\,95.4\,/\,98.3 \\

        \rowcolor{gray!10}
   \cmark
        & \cmark
        & 98.8
        & \textbf{98.5}
        & \textbf{73.9}\,/\,\textbf{94.4}\,/\,\textbf{97.5}
        & \textbf{86.6}\,/\,\textbf{98.8}\,/\,\textbf{98.8} \\

        \bottomrule
    \end{tabular}}
    \vspace{-5pt}
\end{table}

%% file: ablation/filtering.tex
\begin{table}[t]
    \centering
    \caption{{Comparison of view filtering cues and fusion strategies.}}
    \vspace{-5pt}
    \label{tab:cue_ablation}

    \resizebox{\columnwidth}{!}{
    \begin{tabular}{cc|c|cc|cc}
        \toprule
        \multicolumn{2}{c|}{Filtering Cue}
        & \multirow{2}{*}{Fusion}
        & \multirow{2}{*}{CK$\uparrow$}
        & \multirow{2}{*}{DR$\uparrow$}
        & \multirow{2}{*}{T@2/5/10\%$\uparrow$}
        & \multirow{2}{*}{R@2/5/10$^\circ\uparrow$} \\
        DINOv2
        & VGGT Feat.
        &
        &
        &
        &  \\
        \midrule

        \rowcolor{gray!20}
        \multicolumn{7}{c}{\textbf{Phototourism}} \\

        \cmark
        & \xmark
        & --
        & \underline{98.9}
        & \textbf{100.0}
        & \underline{52.4}\,/\,\underline{81.1}\,/\,\textbf{94.2}
        & \underline{96.2}\,/\,\underline{98.9}\,/\,\underline{98.9}
\\

        \xmark
        & \cmark
        & --
        & 91.7
        & \underline{99.7}
        & 47.2\,/\,75.3\,/\,85.4
        & 84.4\,/\,91.7\,/\,91.7
\\

        \cmark
        & \cmark
        & OR
        & 85.0
        & \textbf{100.0}
        & 42.9\,/\,70.0\,/\,80.3
        & 77.8\,/\,85.0\,/\,85.0
\\

        \cmark
        & \cmark
        & Avg
        & 62.8
        & \textbf{100.0}
        & 40.4\,/\,56.7\,/\,62.2
        & 56.1\,/\,62.8\,/\,62.8
\\

        \rowcolor{gray!10}
        \cmark
        & \cmark
        & \textbf{AND}
        & \textbf{100.0}
        & \underline{99.7}
        & \textbf{53.2}\,/\,\textbf{82.0}\,/\,\underline{93.8}
        & \textbf{97.3}\,/\,\textbf{100.0}\,/\,\textbf{100.0}
\\

        \specialrule{1.0pt}{2pt}{2pt}

        \rowcolor{gray!20}
        \multicolumn{7}{c}{\textbf{ETH3D}} \\

        \cmark
        & \xmark
        & --
        & 75.6
        & 95.8
        & 18.6\,/\,40.0\,/\,51.2
        & 46.7\,/\,54.8\,/\,54.8
\\

        \xmark
        & \cmark
        & --
        & \underline{82.1}
        & \underline{99.7}
        & 26.3\,/\,50.6\,/\,\textbf{69.6}
        & \textbf{70.7}\,/\,\textbf{75.0}\,/\,\textbf{75.0}
\\

        \cmark
        & \cmark
        & OR
        & 64.4
        & \textbf{100.0}
        & \textbf{35.0}\,/\,\textbf{58.8}\,/\,\underline{63.2}
        & \underline{60.8}\,/\,63.1\,/\,63.1
\\

        \cmark
        & \cmark
        & Avg
        & 61.9
        & \textbf{100.0}
        & \underline{33.9}\,/\,\underline{53.0}\,/\,57.7
        & 57.7\,/\,57.7\,/\,57.7
\\

        \rowcolor{gray!10}
        \cmark
        & \cmark
        & \textbf{AND}
        & \textbf{99.4}
        & 95.8
        & 14.4\,/\,34.9\,/\,57.1
        & 51.3\,/\,\underline{70.8}\,/\,\underline{73.8}
\\

        \specialrule{1.0pt}{2pt}{2pt}

        \rowcolor{gray!20}
        \multicolumn{7}{c}{\textbf{On-the-Go}} \\

        \cmark
        & \xmark
        & --
        & \textbf{98.8}
        & 98.3
        & \textbf{74.3}\,/\,\textbf{95.8}\,/\,\textbf{97.9}
        & \underline{86.3}\,/\,\textbf{98.8}\,/\,\textbf{98.8}
\\

        \xmark
        & \cmark
        & --
        & 87.1
        & \underline{99.8}
        & 67.8\,/\,82.8\,/\,85.8
        & 84.1\,/\,87.1\,/\,87.1
\\

        \cmark
        & \cmark
        & OR
        & \underline{87.9}
        & \textbf{100.0}
        & 68.8\,/\,84.3\,/\,87.1
        & 83.7\,/\,\underline{87.9}\,/\,\underline{87.9}
\\

        \cmark
        & \cmark
        & Avg
        & 71.3
        & \textbf{100.0}
        & 54.1\,/\,68.1\,/\,70.4
        & 64.2\,/\,71.3\,/\,71.3
\\

        \rowcolor{gray!10}
        \cmark
        & \cmark
        & \textbf{AND}
        & \textbf{98.8}
        & 98.5
        & \underline{73.9}\,/\,\underline{94.4}\,/\,\underline{97.5}
        & \textbf{86.6}\,/\,\textbf{98.8}\,/\,\textbf{98.8}
\\

        \bottomrule
    \end{tabular}}
    \vspace{-5pt}
\end{table}

%% file: ref.bib
@inproceedings{Berton_2025_MegaLoc,
    author    = {Berton, Gabriele and Masone, Carlo},
    title     = {MegaLoc: One Retrieval to Place Them All},
    booktitle = {IEEE/CVF Conference on Computer Vision and Pattern Recognition Workshops},
    month     = {June},
    year      = {2025},
    pages     = {2886--2892}
}

@inproceedings{wang2025vggt,
  title={VGGT: Visual Geometry Grounded Transformer},
  author={Wang, Jianyuan and Chen, Minghao and Karaev, Nikita and Vedaldi, Andrea and Rupprecht, Christian and Novotny, David},
  booktitle={Proceedings of the IEEE/CVF Conference on Computer Vision and Pattern Recognition},
  year={2025}
}

@misc{simeoni2025dinov3,
  title={{DINOv3}},
  author={Sim{\'e}oni, Oriane and Vo, Huy V. and Seitzer, Maximilian and Baldassarre, Federico and Oquab, Maxime and Jose, Cijo and Khalidov, Vasil and Szafraniec, Marc and Yi, Seungeun and Ramamonjisoa, Micha{\"e}l and Massa, Francisco and Haziza, Daniel and Wehrstedt, Luca and Wang, Jianyuan and Darcet, Timoth{\'e}e and Moutakanni, Th{\'e}o and Sentana, Leonel and Roberts, Claire and Vedaldi, Andrea and Tolan, Jamie and Brandt, John and Couprie, Camille and Mairal, Julien and J{\'e}gou, Herv{\'e} and Labatut, Patrick and Bojanowski, Piotr},
  year={2025},
  eprint={2508.10104},
  archivePrefix={arXiv},
  primaryClass={cs.CV},
  url={https://arxiv.org/abs/2508.10104},
}

@inproceedings{SSS:2006,
    author = {Noah Snavely and Steven M. Seitz and Richard Szeliski},
    title = {Photo tourism: Exploring photo collections in 3D},
    booktitle = {SIGGRAPH Conference Proceedings},
    year = {2006},
    isbn = {1-59593-364-6},
    pages = {835--846},
    publisher = {ACM Press},
    address = {New York, NY, USA},
}

@inproceedings{schops2017eth3d,
  title     = {A Multi-View Stereo Benchmark with High-Resolution Images and Multi-Camera Videos},
  author    = {Schops, Thomas and Schonberger, Johannes L. and Galliani, Silvano and Sattler, Torsten and Schindler, Konrad and Pollefeys, Marc and Geiger, Andreas},
  booktitle = {Proceedings of the IEEE Conference on Computer Vision and Pattern Recognition},
  year      = {2017},
  pages     = {3260--3269},
  doi       = {10.1109/CVPR.2017.272}
}

@InProceedings{Ren2024NeRF,
    title={NeRF on-the-go: Exploiting Uncertainty for Distractor-free NeRFs in the Wild},
    author={Ren, Weining and Zhu, Zihan and Sun, Boyang and Chen, Jiaqi and Pollefeys, Marc and Peng, Songyou},
    booktitle={IEEE/CVF Conference on Computer Vision and Pattern Recognition (CVPR)},
    year={2024}
}

@inproceedings{han2026emergent,
  title={Emergent Outlier View Rejection in Visual Geometry Grounded Transformers},
  author={Han, Jisang and Hong, Sunghwan and Jung, Jaewoo and Jang, Wooseok and An, Honggyu and Wang, Qianqian and Kim, Seungryong and Feng, Chen},
  booktitle={Proceedings of the IEEE/CVF Conference on Computer Vision and Pattern Recognition},
  year={2026},
  pages={427--437}
}

@inproceedings{wang2026pi3,
  title={{$\pi^3$: Permutation-Equivariant Visual Geometry Learning}},
  author={Wang, Yifan and Zhou, Jianjun and Zhu, Haoyi and Chang, Wenzheng and Zhou, Yang and Li, Zizun and Chen, Junyi and Pang, Jiangmiao and Shen, Chunhua and He, Tong},
  booktitle={International Conference on Learning Representations},
  year={2026}
}

@article{oquab2024dinov2,
  title={DINOv2: Learning Robust Visual Features without Supervision},
  author={Oquab, Maxime and Darcet, Timoth{\'e}e and Moutakanni, Th{\'e}o and Vo, Huy V. and Szafraniec, Marc and Khalidov, Vasil and Fernandez, Pierre and Haziza, Daniel and Massa, Francisco and El-Nouby, Alaaeldin and Assran, Mido and Ballas, Nicolas and Galuba, Wojciech and Howes, Russell and Huang, Po-Yao and Li, Shang-Wen and Misra, Ishan and Rabbat, Michael and Sharma, Vasu and Synnaeve, Gabriel and Xu, Hu and J{\'e}gou, Herv{\'e} and Mairal, Julien and Labatut, Patrick and Joulin, Armand and Bojanowski, Piotr},
  journal={Transactions on Machine Learning Research},
  year={2024}
}

@inproceedings{wang2024dust3r,
  title     = {DUSt3R: Geometric 3D Vision Made Easy},
  author    = {Wang, Shuzhe and Leroy, Vincent and Cabon, Yohann and Chidlovskii, Boris and Revaud, Jerome},
  booktitle = {Proceedings of the IEEE/CVF Conference on Computer Vision and Pattern Recognition},
  year      = {2024},
  pages     = {20697--20709},
  doi       = {10.1109/CVPR52733.2024.01956}
}

@inproceedings{leroy2024mast3r,
  title     = {Grounding Image Matching in 3D with MASt3R},
  author    = {Leroy, Vincent and Cabon, Yohann and Revaud, Jerome},
  booktitle = {Proceedings of the European Conference on Computer Vision},
  year      = {2024},
  pages     = {71--91},
  doi       = {10.1007/978-3-031-73220-1_5}
}

@inproceedings{duisterhof2025mast3rsfm,
  title     = {MASt3R-SfM: A Fully-Integrated Solution for Unconstrained Structure-from-Motion},
  author    = {Duisterhof, Bardienus Pieter and Zust, Lojze and Weinzaepfel, Philippe and Leroy, Vincent and Cabon, Yohann and Revaud, Jerome},
  booktitle = {International Conference on 3D Vision},
  year      = {2025},
  pages     = {1--10},
  doi       = {10.1109/3DV66043.2025.00008}
}

@inproceedings{wang2024vggsfm,
  title     = {VGGSfM: Visual Geometry Grounded Deep Structure From Motion},
  author    = {Wang, Jianyuan and Karaev, Nikita and Rupprecht, Christian and Novotny, David},
  booktitle = {Proceedings of the IEEE/CVF Conference on Computer Vision and Pattern Recognition},
  year      = {2024},
  pages     = {21686--21697}
}

@article{umeyama1991least,
  title     = {Least-Squares Estimation of Transformation Parameters Between Two Point Patterns},
  author    = {Umeyama, Shinji},
  journal   = {IEEE Transactions on Pattern Analysis and Machine Intelligence},
  volume    = {13},
  number    = {4},
  pages     = {376--380},
  year      = {1991},
  doi       = {10.1109/34.88573}
}

@article{otsu1979threshold,
  title     = {A Threshold Selection Method from Gray-Level Histograms},
  author    = {Otsu, Nobuyuki},
  journal   = {IEEE Transactions on Systems, Man, and Cybernetics},
  volume    = {9},
  number    = {1},
  pages     = {62--66},
  year      = {1979},
  doi       = {10.1109/TSMC.1979.4310076}
}

@InProceedings{yang2025fast3r,
    author    = {Yang, Jianing and Sax, Alexander and Liang, Kevin J. and Henaff, Mikael and Tang, Hao and Cao, Ang and Chai, Joyce and Meier, Franziska and Feiszli, Matt},
    title     = {Fast3R: Towards 3D Reconstruction of 1000+ Images in One Forward Pass},
    booktitle = {Proceedings of the IEEE/CVF Conference on Computer Vision and Pattern Recognition (CVPR)},
    month     = {June},
    year      = {2025},
    pages     = {21924--21935}
}

@InProceedings{cabon2025must3r,
    author    = {Cabon, Yohann and Stoffl, Lucas and Antsfeld, Leonid and Csurka, Gabriela and Chidlovskii, Boris and Revaud, Jerome and Leroy, Vincent},
    title     = {MUSt3R: Multi-view Network for Stereo 3D Reconstruction},
    booktitle = {Proceedings of the IEEE/CVF Conference on Computer Vision and Pattern Recognition (CVPR)},
    month     = {June},
    year      = {2025},
    pages     = {1050--1060}
}

@InProceedings{wang2025cut3r,
    author    = {Wang, Qianqian and Zhang, Yifei and Holynski, Aleksander and Efros, Alexei A. and Kanazawa, Angjoo},
    title     = {Continuous 3D Perception Model with Persistent State},
    booktitle = {Proceedings of the IEEE/CVF Conference on Computer Vision and Pattern Recognition (CVPR)},
    month     = {June},
    year      = {2025},
    pages     = {10510--10522}
}

@InProceedings{wang2026amb3r,
    author    = {Wang, Hengyi and Agapito, Lourdes},
    title     = {AMB3R: Accurate Feed-forward Metric-scale 3D Reconstruction with Backend},
    booktitle = {Proceedings of the IEEE/CVF Conference on Computer Vision and Pattern Recognition (CVPR)},
    month     = {June},
    year      = {2026},
    pages     = {14612--14625}
}

@inproceedings{pan2024glomap,
    author    = {Pan, Linfei and Bar{\'a}th, D{\'a}niel and Pollefeys, Marc and Sch{\"o}nberger, Johannes Lutz},
    title     = {Global Structure-from-Motion Revisited},
    booktitle = {European Conference on Computer Vision (ECCV)},
    year      = {2024}
}

@InProceedings{he2024detectorfree,
    author    = {He, Xingyi and Sun, Jiaming and Wang, Yifan and Peng, Sida and Huang, Qixing and Bao, Hujun and Zhou, Xiaowei},
    title     = {Detector-Free Structure from Motion},
    booktitle = {Proceedings of the IEEE/CVF Conference on Computer Vision and Pattern Recognition (CVPR)},
    month     = {June},
    year      = {2024},
    pages     = {21594--21603}
}

@InProceedings{lee2025densesfm,
    author    = {Lee, JongMin and Yoo, Sungjoo},
    title     = {Dense-SfM: Structure from Motion with Dense Consistent Matching},
    booktitle = {Proceedings of the IEEE/CVF Conference on Computer Vision and Pattern Recognition (CVPR)},
    month     = {June},
    year      = {2025},
    pages     = {6404--6414}
}

@InProceedings{Schonberger_2016_CVPR,
    author = {Schonberger, Johannes L. and Frahm, Jan-Michael},
    title = {Structure-From-Motion Revisited},
    booktitle = {Proceedings of the IEEE Conference on Computer Vision and Pattern Recognition (CVPR)},
    month = {June},
    year = {2016}
}

@article{tang2024mv,
  title={MV-DUSt3R+: Single-Stage Scene Reconstruction from Sparse Views In 2 Seconds},
  author={Tang, Zhenggang and Fan, Yuchen and Wang, Dilin and Xu, Hongyu and Ranjan, Rakesh and Schwing, Alexander and Yan, Zhicheng},
  journal={arXiv preprint arXiv:2412.06974},
  year={2024}
}

@InProceedings{Lindenberger_2021_ICCV,
    author    = {Lindenberger, Philipp and Sarlin, Paul-Edouard and Larsson, Viktor and Pollefeys, Marc},
    title     = {Pixel-Perfect Structure-From-Motion With Featuremetric Refinement},
    booktitle = {Proceedings of the IEEE/CVF International Conference on Computer Vision (ICCV)},
    month     = {October},
    year      = {2021},
    pages     = {5987-5997}
}

@InProceedings{Barath_2020_CVPR,
    author    = {Barath, Daniel and Noskova, Jana and Ivashechkin, Maksym and Matas, Jiri},
    title     = {MAGSAC++, a Fast, Reliable and Accurate Robust Estimator},
    booktitle = {Proceedings of the IEEE/CVF Conference on Computer Vision and Pattern Recognition (CVPR)},
    month     = {June},
    year      = {2020}
}

@InProceedings{Arandjelovic_2016_CVPR,
    author    = {Arandjelovic, Relja and Gronat, Petr and Torii, Akihiko and Pajdla, Tomas and Sivic, Josef},
    title     = {NetVLAD: CNN Architecture for Weakly Supervised Place Recognition},
    booktitle = {Proceedings of the IEEE Conference on Computer Vision and Pattern Recognition (CVPR)},
    month     = {June},
    year      = {2016}
}

@InProceedings{Izquierdo_2024_CVPR,
    author    = {Izquierdo, Sergio and Civera, Javier},
    title     = {Optimal Transport Aggregation for Visual Place Recognition},
    booktitle = {Proceedings of the IEEE/CVF Conference on Computer Vision and Pattern Recognition (CVPR)},
    month     = {June},
    year      = {2024},
    pages     = {17658-17668}
}

@INPROCEEDINGS{9879390,
  author={Lee, Seongwon and Seong, Hongje and Lee, Suhyeon and Kim, Euntai},
  booktitle={2022 IEEE/CVF Conference on Computer Vision and Pattern Recognition (CVPR)}, 
  title={Correlation Verification for Image Retrieval}, 
  year={2022},
  volume={},
  number={},
  pages={5364-5374},
  doi={10.1109/CVPR52688.2022.00530}}

@INPROCEEDINGS{10204303,
  author={Lee, Seongwon and Lee, Suhyeon and Seong, Hongje and Kim, Euntai},
  booktitle={2023 IEEE/CVF Conference on Computer Vision and Pattern Recognition (CVPR)}, 
  title={Revisiting Self-Similarity: Structural Embedding for Image Retrieval}, 
  year={2023},
  volume={},
  number={},
  pages={23412-23421},
  doi={10.1109/CVPR52729.2023.02242}}
